\documentclass{article}

\usepackage{pgfplotstable}
\usepackage{arxiv}
\usepackage{upgreek}

\usepackage[style=alphabetic,natbib=true,maxbibnames=6]{biblatex}

\usepackage{microtype}
\usepackage{graphicx}
\usepackage{subfigure}
\usepackage{booktabs} % for professional tables
\usepackage{siunitx}

\usepackage{algorithm}
\usepackage{algpseudocode}
\usepackage{filecontents}

\DeclareCiteCommand{\citeauthor}
  {\boolfalse{citetracker}%
   \boolfalse{pagetracker}%
   \usebibmacro{prenote}}
  {\ifciteindex
     {\indexnames{labelname}}
     {}%
   \printtext[bibhyperref]{\printnames{labelname}}}
  {\multicitedelim}
  {\usebibmacro{postnote}}

\DeclareCiteCommand{\citeyear}
  {}
  {\bibhyperref{\printdate}}
  {\multicitedelim}
  {}

\usepackage{hyperref}

\usepackage{url} 
\usepackage{graphicx}
\usepackage{amsmath}
\usepackage{amssymb}
\usepackage{amsthm}
\usepackage{tikz}
\usepackage{mathrsfs}
\usepackage{bm}
\usepackage{verbatim}
\usepackage{setspace}
\usepackage{xcolor}
\usepackage{enumitem}
\usepackage[super]{nth}

\usepackage[utf8]{inputenc}
\usepackage{multirow}
\usepackage{hhline}
\usepackage{colortbl}
\usepackage{diagbox} 
\usepackage{makecell}
\usepackage{tabularx}

\usepackage{setspace}
\usetikzlibrary{arrows}
\usetikzlibrary{arrows.meta}
\usetikzlibrary{calc}
\usetikzlibrary{decorations}
\usetikzlibrary{decorations.markings}
\usetikzlibrary{external}
\usetikzlibrary{patterns}
\usetikzlibrary{positioning}
\usetikzlibrary{shapes}
\usetikzlibrary{tikzmark}
\usetikzlibrary{fit}

\title{Multi-objective Constrained Optimization for Energy Applications via Tree Ensembles}
\def\arraystretch{1.0}
\definecolor{highlight}{HTML}{da291c} % central
\newcommand{\review}[1]{#1}

\author{
    Alexander Thebelt\thanks{Corresponding author} \\
    Imperial College London, \\
    South Kensington, SW7 2AZ, UK. \\
    \texttt{alexander.thebelt18@imperial.ac.uk} \\
    \And
    Calvin Tsay \\
    Imperial College London, \\
    South Kensington, SW7 2AZ, UK. \\
    \texttt{c.tsay@imperial.ac.uk } \\
    \And
    Robert M. Lee \\
    BASF SE, \\
    Ludwigshafen am Rhein, Germany. \\
    \texttt{robert-matthew.lee@basf.com} \\
    \And
    Nathan Sudermann-Merx \\
    Cooperative State University Mannheim, \\
    Mannheim, Germany. \\
    \texttt{nathan-georg.sudermann-merx} \\
    \texttt{@dhbw-mannheim.de} \\
    \And
    David Walz \\
    BASF SE, \\
    Ludwigshafen am Rhein, Germany. \\
    \texttt{david-simon.walz@basf.com} \\
    \And
    Tom Tranter \\
    Electrochemical Innovation Lab, \\
    University College London, \\
    Gower St, London WC1E 6BT, UK. \\
    \texttt{t.tranter@ucl.ac.uk} \\
    \And
    Ruth Misener\ensuremath{^*}\kern-\scriptspace \\
    Imperial College London, \\
    South Kensington, SW7 2AZ, UK. \\
    \texttt{r.misener@imperial.ac.uk } \\
}

\begin{document}

\maketitle

\newpage

\begin{abstract}
Energy systems optimization problems are complex due to strongly non-linear system behavior and multiple competing objectives, e.g. economic gain vs. environmental impact. Moreover, a large number of input variables and different variable types, e.g.\ continuous and categorical, are challenges commonly present in real-world applications. In some cases, proposed optimal solutions need to obey explicit input constraints related to physical properties or safety-critical operating conditions. This paper proposes a novel data-driven strategy using tree ensembles for constrained multi-objective optimization of black-box problems with heterogeneous variable spaces for which underlying system dynamics are either too complex to model or unknown. In an extensive case study comprised of synthetic benchmarks and relevant energy applications we demonstrate the competitive performance and sampling efficiency of the proposed algorithm compared to other state-of-the-art tools, making it a useful all-in-one solution for real-world applications with limited evaluation budgets.
\end{abstract}

% keywords can be removed
\keywords{Gradient boosted trees \and multi-objective optimization \and mixed-integer programming
    \and black-box optimization}

\clearpage

\renewcommand{\arraystretch}{1.5}

\section{Introduction}
Energy systems optimization problems frequently exhibit multiple, competing objectives, such as economic vs environmental considerations \cite{PISTIKOPOULOS2021107252}. 
For instance, minimizing energy consumption, incurred costs, and greenhouse gas emissions are differing goals that can result in significantly different optimal operating patterns in energy-intensive processes \cite{kelley2018demand}. 
Likewise, peak performance and long-term degradation frequently represent conflicting objectives when designing new energy systems, such as lithium ion batteries \cite{liu2017optimizing}. 
In these cases, conventional mathematical optimization cannot locate a single solution that is optimal in terms of all given objectives.
Rather, \textit{multi-objective} optimization must be employed, wherein trade-offs between the competing objectives are explored. 
Specifically, multi-objective optimization seeks to find a set of solutions, called Pareto optimal points, that are optimal in terms of the trade-offs between objectives. 
A desired solution can then be selected from the optimal set based on other considerations. 

A second, distinct challenge common in optimization of energy systems is the complexity of involved mathematical models. 
Energy systems are often described by multi-scale, distributed, and/or nonlinear models that can be computationally expensive to evaluate. 
Furthermore, multiple expensive function evaluations may be required to approximate derivative information. 
These models may also involve categorical input variables, e.g.,\ selecting from available equipment,
    influencing the objective functions but for which we cannot compute derivatives or establish a ranking between
    different categories.
Deterministic (global) optimization of the resulting models is invariably impractical using current technologies, motivating instead the use of data-driven, or \textit{derivative-free}, optimization approaches. 
These approaches treat a complex energy systems model as an input-output ``black box'' model during optimization, or a ``grey box'' model if some model equations are retained \cite{BOUKOUVALA2016701, beykal2018optimal}. 
The predominant approaches for black-box optimization are (i) \textit{evolution-based}, where selection heuristics are used to choose promising inputs for successive generations, or (ii) \textit{surrogate-model-based}, where an internal surrogate model is constructed to approximate the input-output behavior \citep{bhosekar2018surrogateReview}. 

Given the above, several recent research efforts have focused on multi-objective, black-box optimization, with particular emphasis on applications in energy systems. 
Successful applications include optimization of wind-farm layouts \cite{rodrigues2016multi, yin2014multi}, building energy management \cite{delgarm2016multi, mayer2020environmental, yu2015application}, microgrid planning \cite{vergara2015towards, zhou2016multi}, process design \cite{sanaye2010thermal, hajabdollahi2017multi}, etc. 
Evolution-based strategies such as genetic algorithms, particle swarm, etc., are a popular choice for such problems, as Pareto points can simply be approximated using the best points found in previous generations \cite{coello2007evolutionary}. 
Nevertheless, these methods often require many function evaluations, making them ill-suited for energy systems models that are expensive to evaluate. 
Moreover, energy-systems problems are typically subject to important constraints (e.g., safety, regulatory limits), which must be satisfied by optimal/feasible solutions \cite{beykal2018optimal}. 
Dealing with constrained input spaces is non-trivial for evolution-based strategies, and information regarding infeasible points is typically unused. 
Rather, infeasible points can be resampled/repaired heuristically \cite{harada2007constraint}, or constraint violations can be treated using a penalty term or additional objective \cite{runarsson2003constrained}. 

Therefore, we focus on Bayesian optimization (BO) strategies, which can exhibit improved sampling efficiency by constructing internal surrogate models. 
Acquisition functions based on the learned surrogate models are then used in optimization, where promising points to sample are identified. 
The use of deterministic optimization tools here can allow energy-system constraints to be handled seamlessly, incorporating knowledge regarding the feasibility of proposed points. 
Many existing methods, such as ParEGO \cite{knowles2006parego} and TSEMO \cite{bradford2018efficient}, employ Gaussian Processes (GPs) as surrogate models. 
GPs are popular choices for black-box optimization because they naturally quantify uncertainty outside of sampled areas, allowing sampling strategies to balance between exploring regions of high uncertainty and exploiting the regions near the best observed values. 
The conflicting nature of these goals is known as the exploitation/exploration trade-off.
Many GP-based approaches handle categorical variables with one-hot encoding, which introduces a continuous variable bounded between zero and one for each category.
This dramatically enlarges the dimensionality for problems with many categorical variables. 
Recent efforts \citep{manson2021mvmoo} propose using tailor-made kernels for GPs to 
    approach multi-objective problems with categorical variables.

Our recent work \cite{thebelt2021entmoot} introduced \texttt{ENTMOOT}, which instead uses tree-based models as the internal surrogates for Bayesian optimization. 
Tree-based models have several notable advantages: they are well-suited for nonlinear and discontinuous functions, and they naturally support discrete/categorical features. 
However, unlike GPs, they do not provide uncertainty estimates, and gradients are not readily available for optimization; these traits have limited the use of tree models in BO \cite{Shahriari2016BO}. 
\texttt{ENTMOOT} overcomes these limitations by (i) employing a novel distance-based metric as a reliable uncertainty estimate, and (ii) using a mixed-integer formulation for optimization over the tree models. 

In this work, we extend the \texttt{ENTMOOT} framework to constrained, multi-objective black-box optimization, and demonstrate its applicability to real-world energy systems optimization. 
In particular, we introduce a weighted Chebyshev method \cite{knowles2006parego} with tree ensemble surrogate models to efficiently identify Pareto optimal points. 
Furthermore, we show that similarity-based metrics provide robust uncertainty estimates for categorical/discrete inputs. 
The proposed framework is first extensively tested using several well-known synthetic benchmarks, and is later demonstrated on the practically motivated case studies of wind-farm layout optimization and lithium-ion battery design. 
These case studies show that \texttt{ENTMOOT} significantly outperforms the popular state-of-the-art \texttt{NSGA-II} tool, largely due to its sample efficiency, effective handling of categorical features and ability to incorporate mechanistic constraints. 
The paper is structured as follows: we first provide background and review related works on BO, tree models, and multi-objective optimization in Section \ref{sec:background}. 
Section \ref{sec:method} then describes the proposed framework, including acquisition function, optimization formulation, and exploration heuristics, in Sections \ref{sec:acq_func}, \ref{sec:tree_enc}, and \ref{sec:explore}, respectively. 
Finally, we give computational results in Section \ref{sec:results}, including first synthetic benchmarks, then two real-world case studies. 
The latter comprise a popular windfarm layout optimization benchmark, as well as a novel battery design optimization study based on the recent software package PyBaMM. 
Both problems are challenging due to having high-dimensional feature spaces, explicit input constraints, and categorical input variables.

\section{Background}
\label{sec:background}
\subsection{Bayesian Optimization}
    Bayesian optimization (BO) is one of the most popular approaches for derivative-free optimization of 
        black-box functions. 
    There are numerous successful applications, ranging from hyperparameter tuning of machine learning 
        algorithms \citep{snoek2015SBOAnn} to design of engineering systems 
        \citep{forrester2008BayesOpt, mockus1989BayesOpt} and drug development 
        \citep{negoescu2011BayesOpt}. 
    BO takes a black-box function $f:\mathbb{R}^n \to \mathbb{R}^{n_f}$ and determines its minimizer 
        $\bm{x}^*$ according to:
    \begin{subequations}
        \label{eq:intro_bo}
        \begin{flalign}
            \bm{x}^* \in &\underset{\bm{x} \in \mathbb{R}^n} {\text{argmin}} \; f(\bm{x}).
            \label{eq:intro_bo_obj} \\
                \; \; \; \text{s.t.} \; \; 
                & g(\bm{x}) = 0, \label{eq:intro_bo_constr_eq} \\
                & h(\bm{x}) \leq 0. \label{eq:intro_bo_constr_ieq}
        \end{flalign}
    \end{subequations}
    Here, $g(\bm{x})$ and $h(\bm{x})$ are additional constraints on inputs $\bm{x}$, i.e.\ equality and 
        inequality constraints, that may be added based on domain knowledge.
    The parameter $n_f$ gives the number of objectives $f(\bm{x})$ and is $n_f = 1$ for 
        single-objective problems.
    In general, no other information of $f(\bm{x})$, e.g.\ derivatives, is available.
    BO learns a probabilistic surrogate model to predict the behavior of function $f(\bm{x})$ to reveal
        promising regions of the search space.
    An acquisition function $Acq(\bm{x})$ is derived from such surrogate models and also takes into account
        less-explored regions, where we expect high prediction errors, capturing the well-known
        \textit{exploitation}/\textit{exploration} trade-off.
    \review{“Many acquisition functions, e.g.\ probability of improvement, expected improvement and upper
        confidence bound \citep{Shahriari2016BO}, combine the surrogate model mean and predictive variance
        into an easy-to-evaluate mathematical expression. Here, the acquisition function seeks to
        balance the mean, which gives an indication for well-performing configurations, and the variance,
        which exposes new areas of the search space.}
    BO solves problem Equ.~\eqref{eq:intro_bo} by sequentially updating the surrogate model and optimizing $Acq(\bm{x})$ to derive 
        promising black-box inputs $\bm{x}_\text{next}$ according to:
    \begin{subequations}
        \label{eq:intro_bo2}
        \begin{flalign}
            \bm{x}_\text{next} \in &\underset{\bm{x} \in \mathbb{R}^n} {\text{argmin}} \; Acq(\bm{x}).
            \label{eq:mod_bo_obj_acq} \\
                \; \; \; \text{s.t.} \; \; 
                & g(\bm{x}) = 0, \label{eq:intro_bo_constr_acq_eq} \\
                & h(\bm{x}) \leq 0. \label{eq:intro_bo_constr_acq_ieq}
        \end{flalign}
    \end{subequations}
    Sophisticated BO algorithms are particularly useful when evaluating $f(\bm{x})$ is expensive, since
        BO tends to be more sample efficient than other heuristics or grid search.
    The most popular choice for surrogate models in BO is the 
        GP \citep{rasmussen2006GP}, as it naturally handles mean and variance 
        predictions to determine confidence intervals.
    Other approaches rely on Bayesian neural networks \citep{snoek2015SBOAnn}, 
        gradient-boosted trees \citep{thebelt2021entmoot} or random 
        forests \citep{hutter2011SequentialModel} as surrogate models.
    For detailed reviews on BO we refer the reader to 
        \citep{brochu2009BayesOpt, frazier2016BayesOpt, frazier2018tutorial, Shahriari2016BO}. \par

\subsection{Tree Ensembles}
    A popular class of data-driven models that is also being used in BO is tree ensembles, e.g.\ 
        gradient-boosted regression trees (GBRTS) \citep{friedman2002stochastic, Friedman2001GreedyMachine} and 
        random forests \citep{breiman2001random}.
    These models can learn discontinuous and nonlinear response surfaces and naturally support
        categorical features, avoiding one-hot encoding or other reformulations.
    Tree ensembles split the search space into separate regions by defining decision 
        thresholds at every node of a decision tree and assigning branches to it, e.g. two branches
        for binary trees.
    These splitting conditions are evaluated to progress in the decision tree and reveal its active leaf.
    Decision trees are highly effective as an ensemble, where the sum of active leaf weights makes up the 
        prediction.
    \review{Therefore, tree ensembles are nonlinear data-driven models that produce a piece-wise
        constant prediction surface.}
    For GBRTs every decision tree is trained to fix predictive errors of previous trees, leading to a strong
        predictive performance of the combined ensemble, i.e.\ multiple weak learners \textit{boost}
        each other to form a strong ensemble learner.
    \review{Boosting methods tend to have an advantage for data with high-dimensional predictors as empirically
        shown in \citet{buhlmann2003boosting}.} \\
    Incorporating tree-based models as surrogate models into BO is difficult \citep{Shahriari2016BO}, with 
        the main challenges being: (i) finding reliable prediction uncertainty metrics that can be used for 
        exploration in BO, and (ii) effectively optimizing the resulting acquisition function, i.e., solving Equ.~\eqref{eq:intro_bo2}, to propose 
        promising new inputs.
    To overcome the former challenge, \textit{Jackknife} and \textit{infinitesimal Jackknife} 
        \citep{JMLRv15wager14a} aim to provide confidence intervals of random forests based
        on statistical metrics.
    In a different approach, \citet{hutter2011SequentialModel} introduced \texttt{SMAC} as the first BO     
        algorithm that uses tree ensembles, i.e. random forests, as surrogate models and estimating variance empirically based on individual decision tree predictions.
    \texttt{SMAC} has been shown to work well compared to other algorithms on various benchmark
        problems.
    However, \citet{Shahriari2016BO} show that the \texttt{SMAC} uncertainty metric has undesirable
        properties in certain settings, e.g.\ narrow confidence intervals in low data regions and 
        uncertainty peaks when there is large disagreement between individual trees.
    \citet{tim_head_2018_1207017} implement quantile regression 
        \citep{Meinshausen06quantileregression, 10.1257/jep.15.4.143} to define upper and lower 
        confidence bounds of tree model predictions in the software tool \textit{Scikit-Opimize} 
        for both GBRTs and random forests.
    Our recent work \citep{thebelt2021entmoot} first introduced \texttt{ENTMOOT}, which combines discrete tree models
        with a distance-based uncertainty metric that is designed to mimic the behavior of 
        GPs.
    We showed using empirical studies that this distance metric effectively reveals areas of 
        high and low model uncertainty. \par
    The other aforementioned challenge of optimizing acquisition functions comprised of tree ensembles 
        is widely ignored, and stochastic strategies, e.g.\ genetic algorithms or random search, are used
        to solved Equ.~\eqref{eq:intro_bo2}.
    To this end, we proposed \citep{thebelt2021entmoot} a global optimization strategy for various distance
        metrics, e.g.\ Manhattan and squared Euclidean distance, based on a tree-model encoding proposed 
        by \citet{Misic2017OptimizationEnsembles} to handle limitations related to stochastic optimization.
    The advantages of global optimization strategies were shown to be especially significant in high-dimensional
        settings, where sampling-based optimization approaches require exponentially many evaluations.
    Moreover, such global optimization strategies guarantee feasibility of constraints on input
        variables and avoid previously mentioned challenges related to missing gradients when it comes to 
        discrete tree models.
    This paper extends the \texttt{ENTMOOT} framework \citep{thebelt2021entmoot} for multi-objective problems and presents
        challenging and relevant benchmarks that further highlight its merits.

\subsection{Multi-Objective Optimization}
    \begin{figure}
        \centering
        \includegraphics[width=0.5\paperwidth]{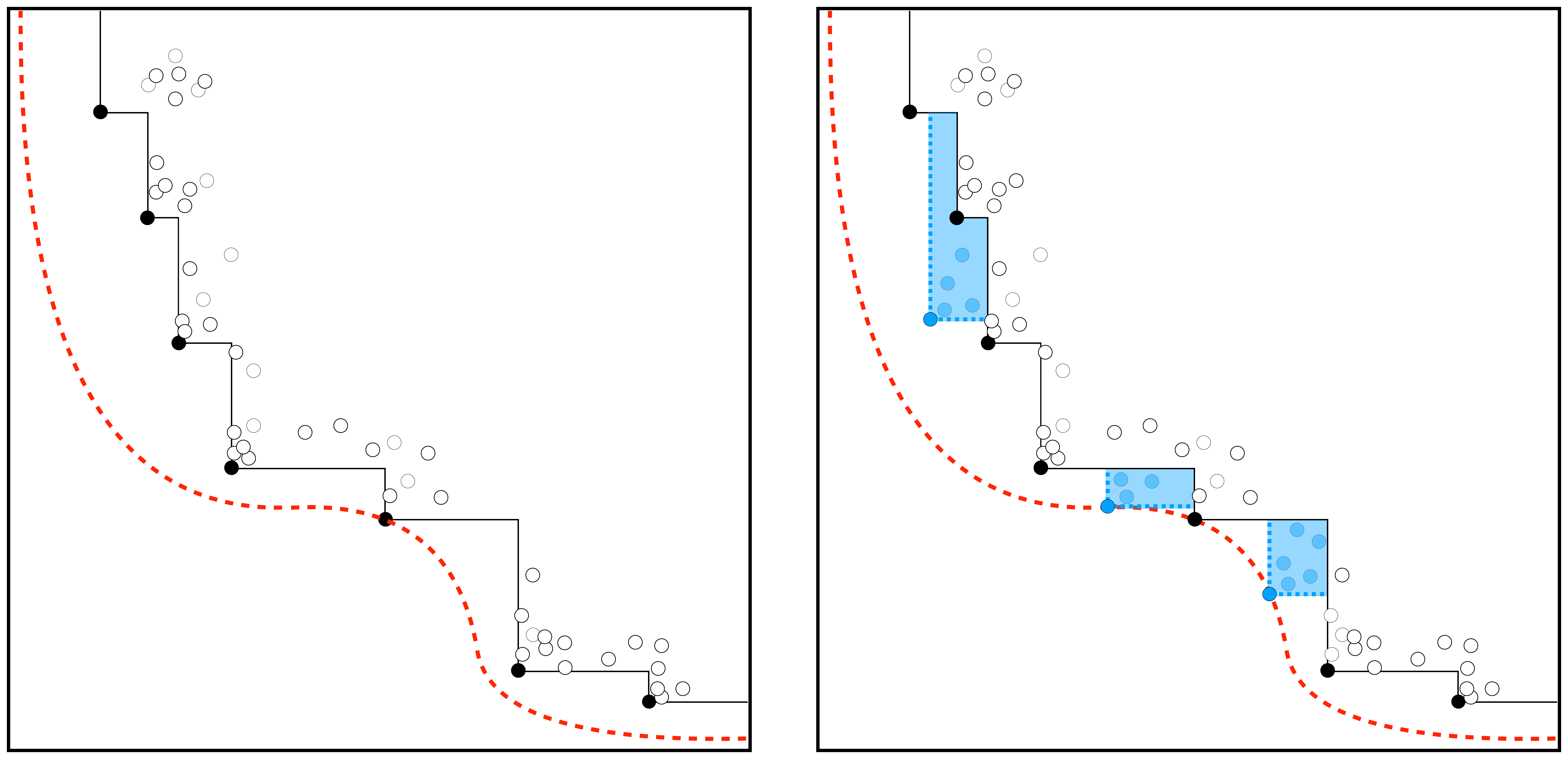}
        \caption{\textbf{(left)}~shows true Pareto frontier (red dashed line) and its approximation (black line) based on non-dominated points (black dots),
            \textbf{(right)}~shows updated approximated Pareto frontier based on new non-dominated points (blue dots).}
        \label{fig:ex_pf}
    \end{figure}
    
    Multi-objective optimization (MOO) problems seek to minimize multiple conflicting objectives 
        $f_i:\mathbb{R}^n \to \mathbb{R}, \forall i \in \left[n_f\right]$.
    Two fundamentally different approaches \citep{hwang2012multiple} in MOO that deal with this problem 
        are scalarization and the Pareto method, which quantify multi-objective trade-offs 
        \textit{a priori} and \textit{a posteriori}, respectively.
    The weighted sum approach \citep{zadeh1963optimality} defines the acquisition function $Acq = \sum\nolimits_{i} w_i f_i(\bm{x})$, representing a 
        scalarization method that combines all objectives into a single objective that is then minimized.
    This strategy is straightforward to implement and has been applied in other constrained black-box optimization
        methods \citep{beykal2018optimal}.
    However, the weighted sum method requires prior knowledge to determine weights $w_i$ for different 
        objectives and offers very limited insights into the trade-offs among various competing objectives. \par
    The Pareto method \citep{pareto1919manuale} seeks to overcome these limitations by exploring the entire
        Pareto frontier comprised by a set of Pareto-optimal points.
    By definition, Pareto-optimal points cannot be improved for one objective function without impairing
        another one.
    Computing the entire Pareto frontier is expensive and may be infeasible due to limited 
        evaluation budgets.
    A common way to approximate Pareto frontiers is to instead use ``non-dominated'' function observations. 
    To give an example, target observation $\bm{Y}_1$ is dominated by $\bm{Y}_2$ if 
        $Y_{1,j} \geq Y_{2,j}, \forall j \in \{1,2, \dots, n_f\}$ and $Y_{1,j} > Y_{2,j}$ for at 
        least one $j \in \{1,2, \dots, n_f\}$.
    Fig.~\ref{fig:ex_pf} shows an example of a two-dimensional Pareto frontier, i.e.\ when both objectives are being 
        minimized, and its approximated Pareto frontier for a given set of data points and when
        more data points are added.
    \begin{figure}
        \centering
        \includegraphics[width=0.75\paperwidth]{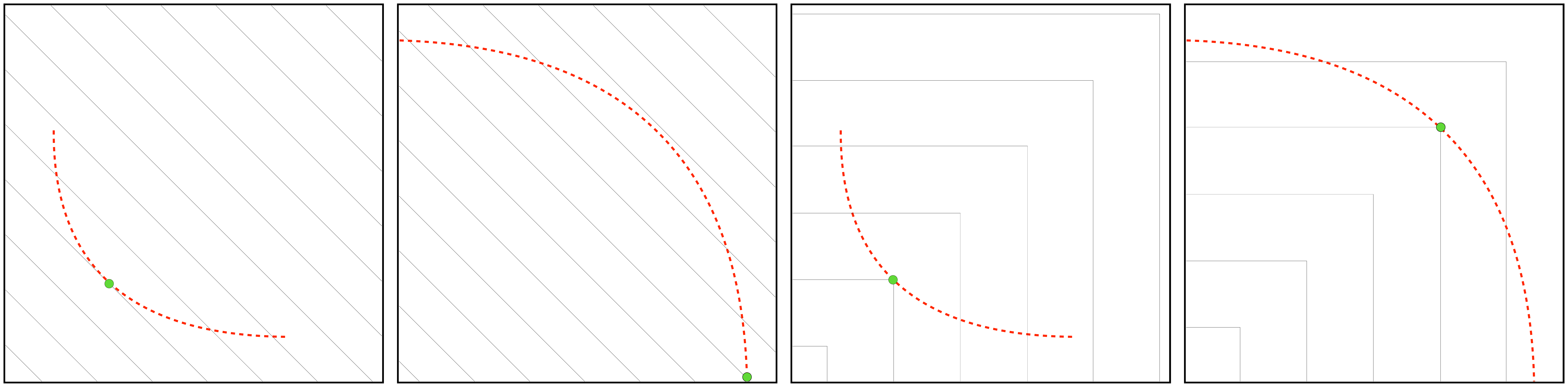}
        \caption{Depicts multi-objective function plot $f_2(f_1(\bm{x}))$ with Pareto frontier (red line), 
        isocontours of $w = (0.5, 0.5)$ scalarization (black lines) and best attainable trade-off based 
        on method used (green dot). \textbf{(left)}~convex Pareto
        frontier using weighted sum approach
            \textbf{(middle-left)}~concave Pareto frontier using weighted sum approach 
            \textbf{(middle-right)}~convex Pareto frontier using Chebyshev method 
            \textbf{(right)}~concave Pareto frontier using Chebyshev method.}
        \label{fig:ex_ws_cs}
    \end{figure}
    The previously mentioned weighted sum approach can generate certain Pareto frontiers by 
        minimizing the objective functions with varying weights $w_i$.
    However, even in the best case, weighted sums can only reconstruct convex 
        Pareto frontiers \citep{messac2000ability}.
    Fig.~\ref{fig:ex_ws_cs} visualizes the Pareto method using the weighted sum objective for convex 
        and concave Pareto frontiers and shows the best attainable points. \par
    An alternative approach implementing the Pareto method is the weighted Chebyshev method 
        \citep{knowles2006parego}, which defines the objective function as 
        $Acq = \text{max}_i \; w_i f_i $.
    % rev2_10
    The weights $w_i$ are drawn from a random distribution and defined such that $\sum \nolimits_i w_i = 1$.
    \review{The choice of distribution can influence what weight combinations the method is
        focusing on, e.g.\ using a normal distribution would focus on more equally distributed
        weights around 0.5. Solving the bilevel optimization problem, i.e.\ $
        \text{min} \; Acq = \text{min} \; \text{max}_i \; w_i f_i $, allows reconstruction of
        Pareto-optimal points on both convex and concave Pareto frontiers.
    This is visualized in Fig.~\ref{fig:ex_ws_cs} where we compare the weighted sum and weighted Chebyshev method.}
    The algorithm proposed here utilizes the weighted Chebyshev method in combination with tree ensembles
        to reveal Pareto-optimal trade-offs of constrained multi-objective optimization problems.
    \review{The $\epsilon$-constraint method is an alternative way of considering different objectives and
        has been applied to energy-related problems \citep{javadi2020multi}.
    This method minimizes one objective while constraining other objectives to a certain range to
        generate the Pareto frontier.
    It is highly effective when there is existing knowledge about some objectives that can be used
        to determine relevant $\epsilon$ values to constrain these objectives.
    However, insufficient $\epsilon$ values may produce non-dominated points far away from the
        true Pareto frontier.} \par
    \review{Other approaches that combine prior knowledge of the problem at hand with different
        data-driven model architectures to determine multi-objective
        trade-offs are given in \citep{olofsson2018bayesian,beykal2018optimal}.}
    Another popular class of methods tackling these problems is genetic algorithms (GA) 
        \citep{kumar2010genetic}.
    One of the most popular algorithms of this category is \texttt{NSGA-II} \citep{deb2002fast}, which handles constrained multi-objective problems and has been successfully deployed in many energy applications \cite{hajabdollahi2017multi, mayer2020environmental, sanaye2010thermal, vergara2015towards, yin2014multi, zhou2016multi, haddadian2017multi, hu2016nsga}.
    While there is no feasibility guarantee for input constraints, \texttt{NSGA-II} generates solutions 
        that seek to minimize constraint violation.
    As a state-of-the-art tool with a readily available Python implementation \citep{pymoo} that can
        handle constrained multi-objective optimization problems, we use it as the main method for 
        comparison benchmarks.

\section{Method}
\label{sec:method}
\begin{table}[]
    \centering
    \begin{tabularx}{\textwidth}{ c|X }
        \hline
        $f_i(\cdot)$ & black-box function of objective $i \in \left[n_f\right]$ \\
        $n$ & number of input dimensions \\
        $n_f$ & number of objectives \\
        $\bm{x}$ & variable input vector of size $n$ \\
        $\bm{x}^*$ & optimal input of black-box function $f$ \\
        $\bm{x}_\text{next}$ & next black-box input proposal \\
        $Acq(\cdot)$ & acquisition function optimized to determine $\bm{x}_\text{next}$ \\
        $g(\cdot)$ & input equality constraints \\
        $h(\cdot)$ & input inequality constraints \\
        $w_i$ & weight for objective $i$ \\
        $\hat{\mu}_i(\cdot)$ & prediction of tree ensemble for objective $i$ \\
        $\alpha(\cdot)$ & uncertainty contribution for exploration \\
        $\kappa$ & hyperparameter that weights exploration term in acquisition function \\
        $\bm{X}$ & input data points $\bm{X} = (\bm{X}_1, \bm{X}_2, \dots, \bm{X}_n)$ \\
        $\bm{Y}$ & target data points $\bm{Y} = (\bm{Y}_1, \bm{Y}_2, \dots, \bm{Y}_n)$ \\
        $\mathcal{N}$ & set of continuous variables \\
        $\mathcal{C}$ & set of categorical variables \\
        $\mathcal{D}$ & \review{available data set} \\
        $S_i(\cdot,\cdot)$ & returns similarity of two data points for categorical feature $i$ \\
        $p^2(i,j)$ & measure of probability for categorical feature $i$ to the take the value $j$ \\
        $\mathcal{P}_a$ & approximated Pareto frontier \\
        $\mathcal{P}_\text{true}$ & true Pareto frontier \\
        \hline
    \end{tabularx}
    \caption{General table of notations.}
    \label{tab:general_not}
\end{table}

In the following sections we give more details on the proposed method.
Section~\ref{sec:acq_func} outlines the acquisition function used.
Sections~\ref{sec:tree_enc} and \ref{sec:explore} explain the encoding for both tree ensembles and
    exploration contributions.
Table~\ref{tab:general_not} summarizes notations used in the following sections.

\subsection{Acquisition Function} \label{sec:acq_func}
    We combine the previously mentioned Chebyshev method with the lower confidence bound acquisition function
        \citep{Cox97sdo:a}, i.e.\ one of the most popular BO acquisition functions, defined as:
    \begin{equation}
        Acq(\bm{x}) = \hat{\mu}(\bm{x}) - \kappa \; \alpha(\bm{x}),
    \end{equation}
    where $\hat{\mu}$ denotes the mean prediction of the surrogate model used.
    $\alpha$ determines exploration based on an uncertainty metric with $\kappa$ functioning as a 
        hyperparameter to balance the \textit{exploitation} / \textit{exploration} trade-off.
    The combined acquisition function replaces the exploitation part with the Chebyshev approach and 
        uses separate tree ensembles to approximate individual objectives $f_i$ with $\hat{\mu}_i$:
    % EQUATIONS: lower confidence bound
    \begin{equation}
        \label{eq:multi_objective}
        \bm{x}_{\text{next}} \in \underset{\bm{x}, \bm{z}, \bm{\nu},
            \hat{\mu}, \alpha} {\text{arg min}} \;
            \underset{i \in \left[n_f\right]}{\text{max}} \; \; w_i
            \frac{\hat{\mu}_i (\bm{x}) - \text{min}(\bm{Y}_i)}
            {\text{max}(\bm{Y}_i) - \text{min}(\bm{Y}_i)} -
            \frac{\kappa}{n} \alpha (\bm{x}).
    \end{equation}
    The next black-box evaluation point $\bm{x}_\text{next}$ is the result of the bilevel problem
        stated in Equ.~\eqref{eq:multi_objective}.
    The first term of Equ.~\eqref{eq:multi_objective} comprises the normalized mean function of
        tree model predictions $\hat{\mu}_i$, $i \in \left[n_f\right]$, trained on individual target 
        columns multiplied with randomly generated parameters $w_i \in \left[0, 1\right]$.
    Note that the normalization of $\hat{\mu}_i (\bm{x})$ is only an approximation based on minimum and
        maximum target observations, i.e. $\text{min}(\bm{Y}_i)$ and $\text{max}(\bm{Y}_i)$,
        in data set $\mathcal{D}$.
    This allows for the optimization to focus on different parts of the Pareto front depending on
        which tree model has the largest weight.
    \review{We emphasize that early iterations are likely going to have poorly normalized objectives,
        i.e.\ when the approximated objective bounds are very different to the actual objective bounds.
    The proposed method allows for users to explicitly specify $\text{min}(\bm{Y}_i)$ and
        $\text{max}(\bm{Y}_i)$ , to avoid poor normalizations of objective values if objective
        bounds are known a priori.}
    The second term of Equ.~\eqref{eq:multi_objective} is proposed by \citet{thebelt2021entmoot} and helps     
        exploration  by incentivizing solutions away from previously explored
        data points.
    We reformulate optimization objective in Equ.~\eqref{eq:multi_objective} to remove its bilevel
        nature according to:
    % EQUATIONS: lower confidence bound
    \begin{subequations}
        \label{eq:multi_objective_reform}
        \begin{flalign}
            \bm{x}_{\text{next}} \in & \underset{\bm{x}, \bm{z}, \bm{\nu},
            \hat{\mu}, \alpha} {\text{arg min}} \;
                \hat{\mu} - \frac{\kappa}{n} \alpha (\bm{x}) \\
            & \text{s.t.} \; \hat{\mu} \geq w_i
                \frac{\hat{\mu}_i (\bm{x}) - \text{min}(\bm{Y}_i)}
                {\text{max}(\bm{Y}_i) - \text{min}(\bm{Y}_i)}, \;
                \forall i \in \left[n_f\right]
        \end{flalign}
    \end{subequations}
    Formulations in Equ.~\eqref{eq:multi_objective} and Equ.~\eqref{eq:multi_objective_reform} are
        equivalent since $\hat{\mu}$ must be greater or equal to all individual tree model 
        contributions. Therefore, at the optimal point, it will be equivalent to the largest tree model prediction, making 
        other tree model contributions redundant.

\subsection{Tree Ensemble Encoding} \label{sec:tree_enc}
    \begin{table}[]
        \centering
        \begin{tabularx}{\textwidth}{ c|X }
            \hline
            $z_{t,l}$ & variable leaf $l$ in tree $t$ \review{acting as a binary variable} \\
            $\nu_{i,j}$ & binary variable for active split of feature $i$ \\
            $v_{i,j}$ & numerical value split node $j$ of feature $i$ \\
            $v_i^L$, $v_i^U$ & lower and upper bound of continuous feature $i$ \\
            $F_{t,l}$ & weight of leaf $l$ in tree $t$ \\
            $\mathcal{T}$ & set of trees \\
            $\mathcal{L}_t$ & set of leaves in tree $t$ \\
            $\mathbf{splits}(t)$ & set of splits of tree $t$ \\
            $\mathbf{left}(s)$ & leaf indices left of split $s$ \\
            $\mathbf{right}(s)$ & leaf indices right of split $s$ \\
            $\text{V}(s)$ & feature participating in split $s$ \\
            \hline
        \end{tabularx}
        \caption{Tree ensemble encoding table of notation.}
        \label{tab:tree_model}
    \end{table}
    
    % FIGURE: tree ensemble encoding
    \begin{figure*}
    \label{fig:dec_tree}
        \begin{center}
            \includegraphics[width=0.77\paperwidth]{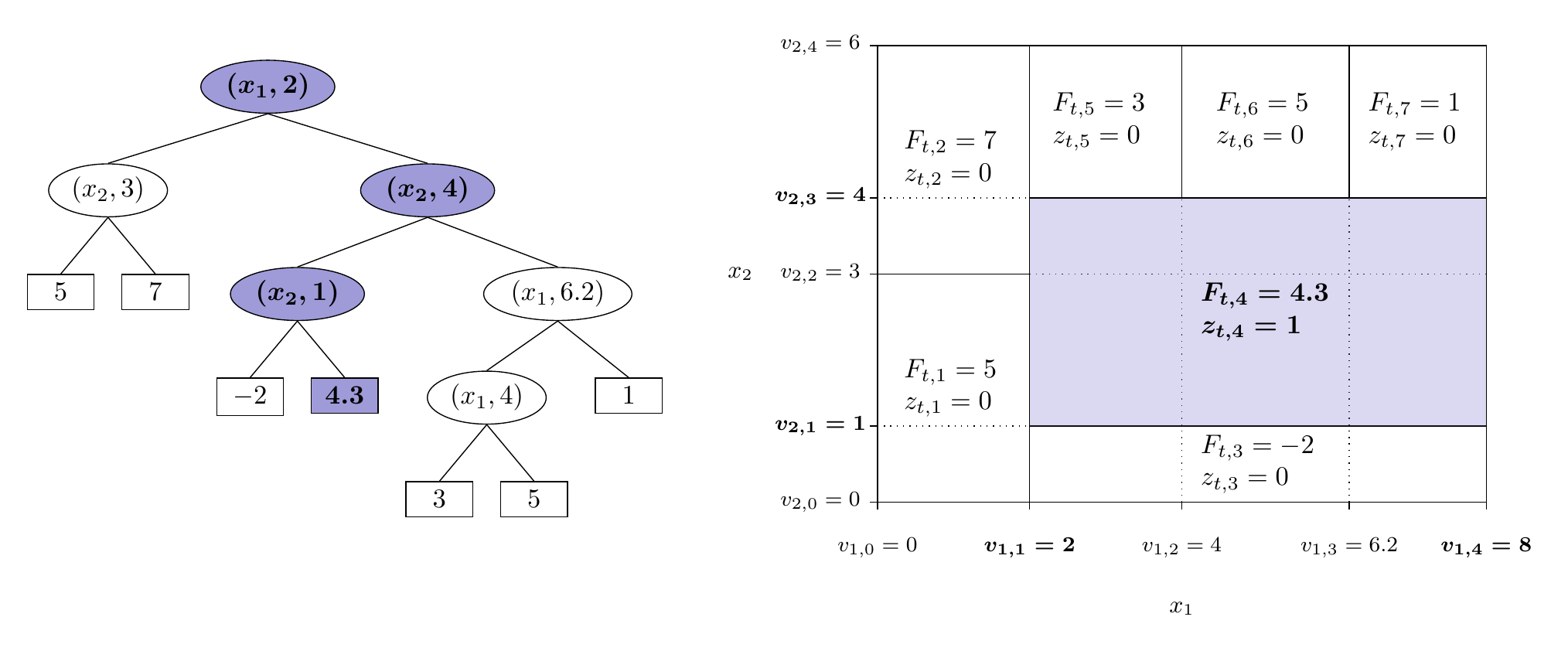}
        \end{center}
        \caption{Visualization of a decision tree evaluation.}
    \end{figure*}
    
    The proposed method uses the \citet{Misic2017OptimizationEnsembles} mixed-integer linear
        programming formulation to encode the tree model \cite{Mistry2018Mixed-IntegerEmbedded}.
    Tree ensembles are a popular choice for problems with different variable types, i.e.\
        continuous and categorical variables.
    Continuous variables are defined based on a lower and upper bound, and individual decision trees
        use splitting conditions to partition the input space.
    For categorical variables we distinguish between nominal and ordinal variables.
    For different categories of ordinal variables we can establish an order, but cannot quantify the
        relative distance between different categories.
    On the other hand, nominal variables neither define an order nor a distance between different
        categories.
    Training a tree ensemble determines split conditions in individual decision trees for different 
        variable types.
    Table~\ref{tab:tree_model} defines the notation in this section.
    For continuous variables trees define numerical thresholds $v_{i,j}$ as split conditions
        that determine whether the decision tree evaluation progresses to the left or right 
        branch of the node.
    The index $i$ indicates the feature involved in the split and $j \in \{1,2,\dots,K_i\}$
        denotes an ordering for all continuous split conditions in the ensemble according to
        $v_{i,1} < v_{i,2} < ... < v_{i,K_i}$.
    To assign the correct numerical value $v_{i,j}$ to a split $s \in \mathbf{splits}(t)$ in tree $t$
        we define $\mathbb{C}(s)=j$ which maps split $s$ onto the correct $j\in 
        \{1,2,\dots,K_{\text{V}(s)}\}$.
    $\text{V}(s)$ gives the index of the feature participating in split $s$. 
    On the other hand, categorical variables define split conditions as subsets of all possible categories
         for a feature $i$.
    LightGBM implements this procedure by grouping for maximum homogeneity as proposed 
        by \citet{fisher1958grouping}.
    \review{With slight abuse of notation, we define $\mathbb{C}(s) \subseteq \{1,2,\dots,K_{\text{V}(s)}\}$
        for categorical features.}
    The set $\{1,2,\dots,K_{\text{V}(s)}\}$ has all available categories if $\text{V}(s)$ is a
        categorical feature.
    Defining the tree ensemble in this way simplifies the encoding as an optimization model (see Table~\ref{tab:tree_model}).
\par
    We start by defining $\mathcal{N}$ and $\mathcal{C}$ as sets for continuous and categorical 
        variables, respectively.
    For continuous variables the model defines binary variables $\nu_{i,j}$ corresponding to split
        condition: $\nu_{i,j} = \mathbb{I} \{ x_i \leq v_{i,j} \},
        \forall i \in \mathcal{N}, j \in \{ 1,...,K_i\}$.
    For categorical variables the model defines binary variables $\nu_{i,j}$ corresponding to split
        condition: $\nu_{i,j} = \mathbb{I} \{ x_i=j \}, \forall i \in \mathcal{C}, j \in \{ 1,...,K_i\}$.
    The following constraints encode the tree model for single objectives according 
        to \cite{Misic2017OptimizationEnsembles}:
    % EQUATIONS: gbrt model
    \begin{subequations}
    \label{eq:tree_encoding}
    \begin{flalign}
        &\hat{\mu} =\sum\limits_{t\in{\mathcal{T}}} \sum\limits_{l\in{\mathcal{L}_{t}}}
            F_{t,l} z_{t,l}, & & \label{eq:const_a}\\
        &\;\;\;\sum\limits_{l\in{\mathcal{L}_{t}}}\;\;\;\;
            z_{t,l} = 1, & &\forall t\in \mathcal{T}, \label{eq:const_b}\\
        &\sum\limits_{\;\;l\in\mathbf{left}(s)}\; z_{t,l} \leq
            \sum\limits_{j \in \mathbf{C}(s)} \nu_{\text{V}(s),j},
            & &\forall t \in \mathcal{T},
            \forall s \in \mathbf{splits}(t), \label{eq:const_c}\\
        &\sum\limits_{\;\;l\in\mathbf{right}(s)} z_{t,l} \leq
            1 - \sum\limits_{j \in \mathbf{C}(s)} \nu_{\text{V}(s),j},
            & &\forall t \in \mathcal{T},
            \forall s \in \mathbf{splits}(t), \label{eq:const_d} \\
        &\;\;\;\sum\limits_{j=1}^{K_i}\;\;\; \nu_{i,j} = 1,
            & &\forall i \in \mathcal{C}, \label{eq:const_e} \\
        &\nu_{i,j} \leq \nu_{i,j+1}, & &\forall i \in \mathcal{N},
            \forall j \in \left [ K_i - 1 \right ], \label{eq:const_f} \\
        &\nu_{i,j} \in \{ 0,1  \}, & &\forall i \in \left [ n \right ],
            \forall j \in \left [ K_i \right ], \label{eq:const_g} \\
        &z_{t,l} \geq 0, & &\forall t\in{\mathcal{T}},
            \forall l\in{\mathcal{L}_{t}}. \label{eq:const_h}
    \end{flalign}
    \end{subequations}
    \review{For every tree ensemble $\hat{\mu}_i$ of Equ.~\eqref{eq:multi_objective_reform},
        we define the constraints Equ.~\eqref{eq:const_a} as the sum of all active leaf weights
        $F_{t,l}$, which are obtained from training of the tree ensemble.}
    The leaves are indexed by $t\in{\mathcal{T}}$ and $l\in{\mathcal{L}_{t}}$, with $\mathcal{T}$ 
        and $\mathcal{L}_t$ denoting the set of trees and the set of leaves in tree $t$, respectively. 
    Binary variables $z_{t,l} \in \left[ 0,1 \right]$ function as switches indicating which leaves
        are active.
    \review{For this model definition \citet{Misic2017OptimizationEnsembles} shows that variables
        $z_{t,l}$ can be relaxed to positive continuous variables according to Equ.~\eqref{eq:const_h}
        without losing their binary behavior, as they are fully-defined by binary variables
        $\nu_{\text{V}(s),j}$, i.e.\ Equ.~\eqref{eq:const_c} and Equ.~\eqref{eq:const_d}.
    Relaxing binary variables can allow third party solvers to use more effective solution methods.}
    Equ.~\eqref{eq:const_b} ensures that only one leaf per tree contributes to the tree ensemble
        prediction. 
    Equ.~\eqref{eq:const_c}, \eqref{eq:const_d} and \eqref{eq:const_f} force all splits 
        $s \in \mathbf{splits}(t)$, leading to an active leaf, to occur in the correct order.
    Here $\mathbf{left}(s)$ and $\mathbf{right}(s)$ denote subsets of leaf indices
        $l \in \mathcal{L}_t$ which are left and right of split $s$ in tree $t$.
    Binary variables $\nu_{\text{V}(s),j}$ determine which splits are active.
    $\mathbf{C}(s)$ behaves differently depending on the variable type, i.e.\ continuous or 
        categorical, as previously explained.
    To ensure that only one of the available categories for feature $i$ is active, Equ.~\eqref{eq:const_e}
         constraints are added for all categorical variables $i \in \mathcal{C}$. 
    Fig.~\ref{fig:dec_tree} visualizes the evaluation of a single decision tree and shows which leaf variables are active.
    For more details on the tree ensemble encoding and theoretical properties of the formulation we 
        refer the reader to \citep{Misic2017OptimizationEnsembles}. \\
    \review{
    Some key hyperparameters of tree ensembles are the number of trees, the maximum depth of a
        single decision tree and the minimum number of data points used to define a leaf.
    While tuning these hyperparameters may improve tree ensemble performance, we assume a pre-defined
        set of fixed hyperparameter values to allow for a fair comparison with other methods.
    Standard techniques, e.g.\ cross-validation or early stopping, could be used to further improve
        \texttt{ENTMOOT}'s performance and help with exploring unknown areas where model
        uncertainty is high.
    } \par
        
\subsection{Search Space Exploration} \label{sec:explore}
    Next we present details about the Equ.~\eqref{eq:multi_objective_reform} exploration term denoted by 
        $\alpha$.
    \review{Given that tree ensembles have a piece-wise constant prediction surface with no
        built-in metric to evaluate uncertainty, we introduce a distance-based exploration measure
        that incentivizes solutions away from already visited data points.}
    Different strategies to quantify uncertainty for continuous and categorical variables are deployed 
        and combined to define $\alpha$ according to:
    % EQUATION: exploration
    \begin{subequations}
        \label{eq:alphas}
    \begin{flalign}
        & \alpha \leq \alpha_{\mathcal{N}}^d + \alpha_{\mathcal{C}}^d,\; \;
            \forall d \in \mathcal{D} \\
        & \alpha_{\mathcal{N}}^d \in \left[ 0, |\mathcal{N}| \right] \\
        & \alpha_{\mathcal{C}}^d \in \left[ 0, |\mathcal{C}| \right]
    \end{flalign}
    \end{subequations}
    with $\alpha_{\mathcal{N}}^d$ and $\alpha_{\mathcal{C}}^d$ denoting exploration contributions of
        continuous and categorical variables, respectively.
    For continuous variables we define $\alpha_{\mathcal{N}}^d$ by introducing distance constraints
        according to:
    % continous vars
    \begin{equation}
        \label{eq:dist_meas}
        \alpha_\mathcal{N}^d \leq  \left\lVert
            \frac{\bm{x}_\mathcal{N} - \bm{v}^{L}}
            {\bm{v}^{U} - \bm{v}^{L}}
            - \frac{\bm{X}_d - \bm{v}^{L}}
            {\bm{v}^{U} - \bm{v}^{L}}
            \right\rVert_2^2,
            \; \; \forall d \in \mathcal{D}
    \end{equation}
    Continuous variable bounds $\bm{v}^{L}$ and $\bm{v}^{U}$ normalize the vector of continuous variables
        $\bm{x}_\mathcal{N}$.
    Equ.~\eqref{eq:dist_meas} defines $\alpha_\mathcal{N}$ as the squared Euclidean distance to the
        closest normalized data point $\bm{X}_d$ and bounds it to $\alpha \in \left[0,1\right]$.
    Due to the negative contribution of $\alpha$ in Equ.~\eqref{eq:multi_objective_reform}, the 
        Equ.~\eqref{eq:dist_meas} quadratic constraints make the overall problem non-convex.
    These types of non-convex quadratic problems can be solved by commercial branch-and-bound solvers, e.g.\
        Gurobi~-~v9.
    The proposed measure is similar to \citet{thebelt2021entmoot} and can be reformulated to
        utilize the standard Euclidean or Manhattan distances.
    To correlate continuous variables $\bm{x}$ with discrete tree model splits, we introduce linking
        constraints according to Equ.~\eqref{eq:linking_const} which assign separate
        intervals $v_{i,j}$ defined by the tree model splits back to the original continuous
        search space $\bm{x} \in \mathbb{R}^n$.
    % EQUATIONS: linking
        Parameters $v_i^L$ and $v_i^U$ denote upper and lower bounds of feature $i$, respectively.
    \begin{subequations}
    \label{eq:linking_const}
    \begin{flalign} 
        &x_{i} \geq v^L_{i} + \sum\limits_{j=1}^{K_{i}} \left (v_{i,j} - v_{i,j-1} \right ) 
            \left ( 1 - \nu_{i,j} \right ), & &\forall i \in \mathcal{N},\\
        &x_{i} \leq v^U_{i} + \sum\limits_{j=1}^{K_{i}} \left (v_{i,j} - v_{i,j+1} \right ) 
            \nu_{i,j}, & &\forall i \in \mathcal{N},\\
        &x_{i} \in \left [ v_{i}^{L},v_{i}^{U} \right ], & &\forall i \in \mathcal{N}.
    \end{flalign}
    \end{subequations}
    Quantifying exploration contributions as distances is more difficult when it comes to
        categorical variables.
    Here we use similarity measures proposed by \citet{boriah2008Categ} that compute the proximity
        of categorical feature values between two points of the data set.
    \review{These measures define $S_i \left(\bm{X}_{d_1,i}, \bm{X}_{d_2,i} \right)$ as the
        similarity between feature $i$ of data points $\bm{X}_{d_1}$ and $\bm{X}_{d_2}$ with regard to
        categorical feature $i \in \mathcal{C}$.}
    The easiest way to define $S_i$ is the \textit{Overlap} measure:
    % EQUATION: overlap
    \begin{equation}
    \label{eq:overlap}
    \review{S_i\left(\bm{X}_{d_1,i}, \bm{X}_{d_2,i}\right)}=
        \begin{cases}
            1 \; \; \bm{X}_{d_1,i} = \bm{X}_{d_2,i}\\
            0 \; \;  \text{otherwise}
        \end{cases}
    \end{equation}
    Using this measure we observe a range $S_i \in \left[ 0, 1 \right]$ since it only measures the overlap 
        between categorical features, i.e.\ having the same category gives similarity of one while having different 
        categorical values give similarity of values.
    While this measure is easy to implement and effective, it does not take into consideration the
        number of times different categorical data points already appear in the data set.
    Here, \citet{boriah2008Categ} propose probability based metrics to consider different types of
        similarities.
    The property $p^2(i,j)$ is a probability estimate of category $j$ of feature $i$ appearing in data
        set $\mathcal{D}$ and is defined as:
    % EQUATION: p^2
    \begin{equation}
        p^2\left(i, j\right) = \frac{\text{count}_i \left( j\right) \left( \text{count}_i
            \left( j\right)-1 \right) }{\mid \mathcal{D} \mid
            \left( \mid \mathcal{D} \mid-1\right)}
    \end{equation}
    Large values of $p^2(i,j)$ indicate a high probability that categorical feature $i$ will have
        the value $j$.
    \citet{boriah2008Categ} use $p^2(i,j)$ to define the \textit{Goodall4} similarity according to
    % EQUATION: goodall4
    \begin{equation}
        \label{eq:goodall4}
        \review{S_i\left(\bm{X}_{d_1,i}, \bm{X}_{d_2,i}\right)}=
            \begin{cases}
                p^2\left(i, \bm{X}_{d_1, i}\right) &\text{if} \;
                    \bm{X}_{d_1, i} = \bm{X}_{d_2, i}\\
                0 &\text{otherwise}
            \end{cases}
    \end{equation}
    For categorical variables we link the distance metric to the exploration term by adding the following 
        constraints to our optimization problem:
    % EQUATION: cat distance
    \begin{equation}
        \label{eq:cat_dist}
        \alpha_\mathcal{C}^d \leq \sum\limits_{i \in \mathcal{C}} \sum\limits_{j = 1}^{K_i}
            \left[ 1 - \review{S_i\left(\bm{X}_{d,i},j\right)} \right] \nu_{i,j}, \; \;
            \forall d \in \mathcal{D}
    \end{equation}
    Equ.~\eqref{eq:goodall4} computes larger similarities for features that occur more often in the
        data set, causing $\alpha_\mathcal{C}^d$ in Equ.~\eqref{eq:cat_dist} to take smaller values,
        which incentivizes solutions that appear less frequently in data set $\mathcal{D}$.
    \review{We can read active categorical variable values by checking which of
        the $\nu_{i,j}$ binary variables is not zero:}
    % EQUATION: cat mapping
    \begin{equation}
        \label{eq:cat_map}
        \review{x_i = \underset{j \in \left [ K_i \right ]}{\text{argmax}} \;
            \{\nu_{i,j}\}, \forall i \in \mathcal{C}.}
    \end{equation}
    \review{We note that Equ.~\eqref{eq:cat_map} is not part of the optimization problem and should
        demonstrate how active labels are linked to $\bm{x}_\text{next}$.}
    In summary, the resulting optimization model that determines the next promising input 
        $\bm{x}_\text{next}$ uses Equ.~\eqref{eq:multi_objective_reform}, 
        Equ.~\eqref{eq:tree_encoding} for every tree model, Equ.~\eqref{eq:dist_meas}, 
        Equ.~\eqref{eq:linking_const} and Equ.~\eqref{eq:cat_dist}, and is classified as a mixed-integer
        non-convex quadratic problem.
\subsection{Combined Model}
\review{
The combined model consists of the Section~\ref{sec:tree_enc} tree ensemble encodings and the
    Section~\ref{sec:explore} exploration measure.
The proposed method uses the modified Chebyshev method introduced in Section~\ref{sec:acq_func}.
Optimization of the tree model encoding is an NP-hard problem \citep{Misic2017OptimizationEnsembles},
    and, together with the exploration measure, can be classified as a nonconvex mixed-integer quadratic program.
The nonconvex part originates from the definition of $\alpha_\mathcal{N}^d$ as a quadratic in
    Equ.~\eqref{eq:dist_meas}, which is then maximized in the objective function, i.e.\ its
    negative value is minimized.
While these formulations do not scale well with growing problem sizes, \citet{thebelt2021entmoot}
    found that most problems converge within minutes due to the usage of effective heuristics in third party solvers.
\citet{Misic2017OptimizationEnsembles}, \citet{thebelt2021entmoot} and \citet{Mistry2018Mixed-IntegerEmbedded}
    also propose custom heuristics specifically tailored for optimization problems including tree
    ensembles to enhance bound improvement when optimizing such models. \\
The tree encoding defined in Equ.~\eqref{eq:tree_encoding} adds constraints for splits in every
    decision tree and scales with the size of the tree ensemble.
While it is independent of the number of data points and the dimensionality, one would expect larger
    tree ensembles for large data sets and high-dimensional spaces, and therefore more constraints in the optimization model.
The uncertainty metric defined in Section~\ref{sec:explore} directly scales with the size of the data set,
    dimensionality as well as number of categories if categorical variables are involved.
Equ.~\eqref{eq:alphas} and Equ.~\eqref{eq:dist_meas} are added for every additional data point, and
    the vectors in Equ.~\eqref{eq:dist_meas} grow with increasing dimensionality, making the resulting
    problem more difficult to solve.
Moreover, similarity matrix $S$ for categorical variables grows with increasing number of data points
    and Equ.~\eqref{eq:cat_dist} is added for every data point.
\citet{thebelt2021entmoot} solve a similar problem for single objective problems and propose using
    data clustering to combine multiple data points into single cluster centers to reduce
    the size of the problem.
While the proposed method is inherently difficult to solve, we motivate its usefulness by empirically
    evaluating it on practically motivated case studies.
The superior performance with respect to sampling efficiency of the proposed method motivates
    its usage over cheaper algorithms, e.g.\texttt{NSGA-II}, especially when black-box functions
    are expensive to evaluate.
}

\section{Numerical Studies}
\label{sec:results}
This section empirically evaluates the framework proposed in this paper.
Code recapitulating the \texttt{ENTMOOT} results can be found here: \url{https://github.com/cog-imperial/moo_trees}.
We select synthetic benchmarks with known Pareto fronts that enable comprehensive
    comparison with respect to performance metrics.
Moreover, we present two practically motivated case studies, i.e.,\ wind turbine placement (WTP)
    and battery material selection (BMS), that emphasize the advantages of the proposed method and
    highlight its relevance to energy systems.
\review{For all studies we assume expensive black-box evaluations comprise the majority of the total
    computational complexity of each run. Therefore, we evaluate solver performance based on how
    many black-box evaluations are needed to explore the Pareto frontier, i.e. different
    metrics are used to measure the progress.}
For all experiments we provide medians for all performance metrics, as well as first and third quartiles
    that bound the confidence intervals, computed using 25 runs with ordered random seeds in $\{101,102,...125\}$.
The same set of initial points are provided for all competing methods to facilitate comparison.
\review{By using the same set of initial data points and testing multiple random seeds,
    we mitigate the effects of specific initial settings favoring one method over the other.}
% rev2_17
\review{For the case studies with unknown Pareto frontiers we compute the bounded hypervolume of
    the objective space over number of black-box evaluations.
The number of black-box evaluations contributes to the overall computational complexity,
    and we therefore compare methods based on how many black-box evaluations are needed
    to improve the bounded hypervolume of the objective space.} \\
We compare the proposed methods to \texttt{NSGA-II} using its \textit{pymoo} \citep{pymoo} implementation.
Additionally, we investigate random search for unconstrained problems using a feasible sampling strategy, as described in
    Section~\ref{sec:wind_opt_bench}.
The \texttt{NSGA-II} method is used with default hyperparameter values in \textit{pymoo}, except for initial population size, which is set consistently with the other considered methods.
Since \texttt{NSGA-II} does not directly support categorical variables, we apply one-hot encoding by introducing continuous
    auxiliary variables for all categories of a categorical variable bounded within $\left[0.0,1.0\right]$.
The auxiliary variable with the highest numerical value determines the category picked.
\texttt{ENTMOOT} uses 400 trees with a maximum tree depth of three and a minimum of two data points per leaf.
The tree ensemble is trained using LightGBM \citep{Ke2017LightGBM:Tree}, and the default value of $\kappa = 1.96$ is picked for 
    all numerical studies.
\review{To allow for a fair comparison with other methods, we used the same set of hyperparameters for
    \texttt{ENTMOOT} throughout all runs.}
% rev2_7
\review{\texttt{ENTMOOT} uses Gurobi~-~v9 to solve the mixed-integer quadratic program (MIQP) with
    default parameter settings to find minima of the tree ensemble surrogate model.
We found that for most instances Gurobi~-~v9 finds a global optimal solution in a fairly short
    time given the default relative optimality gap of \num{1e-4} and feasibility tolerance of \num{1e-6}.
However, sometimes Gurobi~-~v9 struggles to fully close the optimality gap, so we enforce a time
    limit of 100 s, given that a feasible solution is found, due to computational limitations when running batch jobs.
If a feasible solution is not found, the optimization procedure is continued until one is found.
We note that this procedure only affects the method presented here, and not other tools that we compare against.}
All experiments are run on a Linux machine equipped with an Intel Core i7-7700K 4.20 GHz and 16 GB of
    memory.
    
\subsection{Performance Metrics}
We use multiple different measures to compare the performance of different algorithms.
For benchmarks where the true Pareto frontier is known, we rely on the Euclidean generational
    distance (GD) \citep{van1999multiobjective}, the inverted generational distance (IGD) 
    \citep{sierra2004new}, the maximum Pareto frontier error (MPFE) \citep{van1999multiobjective}
    and the volume ratio (VR) \citep{olofsson2018bayesian}.
The true Pareto frontier is denoted as $\mathcal{P}_\text{true}$ and $\mathcal{P}_a$ is the 
    approximated Pareto frontier derived by different algorithms.
For synthetic benchmarks in Section~\ref{sec:synthetic_benchmarks}, $\mathcal{P}_\text{true}$ is determined 
    by running \texttt{NSGA-II} for a few thousand iterations.
For energy systems related benchmarks there is no true Pareto frontier available, and we compare
    the hypervolume bounded by the approximate Pareto frontiers derived from different algorithms.
    
We compute GD according to:
% EQUATION: GD
\begin{equation}
    \text{GD}\left( \mathcal{P}_a \right) = \frac{1}{|\mathcal{P}_a|}
        \sum\limits_{\bm{r} \in \mathcal{P}_a}
        \underset{\bm{r}^{\left( t \right)} \in \mathcal{P}_\text{true}}{\text{min}} \;
        \left\lVert \bm{r} - \bm{r}^{\left( t \right)} \right\rVert_2,
\end{equation}
which describes the average distance of points in in the approximated Pareto frontier $\mathcal{P}_a$ to 
    the true Pareto frontier $\mathcal{P}_\text{true}$.
The results of this measure may be misleading if the approximate number consists of only a few 
    good Pareto points.
To overcome this problem we also report IGD which computes the average distance of points on the true
    Pareto frontier to the approximate Pareto frontier according to:
% EQUATION: IGD
\begin{equation}
    \text{IGD}\left( \mathcal{P}_a \right) = \frac{1}{|\mathcal{P}_\text{true}|}
        \sum\limits_{\bm{r}^{\left( t \right)} \in \mathcal{P}_{\text{true}}}
        \underset{\bm{r} \in \mathcal{P}_a}{\text{min}} \;
        \left\lVert \bm{r}^{\left( t \right)} - \bm{r} \right\rVert_2.
\end{equation}
The MPFE measure is defined as:
% EQUATION: MPFE
\begin{equation}
    \text{MPFE}\left( \mathcal{P}_a \right) =
        \underset{\bm{r}^{\left( t \right)} \in \mathcal{P}_\text{true}}{\text{max}} \;
        \underset{\bm{r} \in \mathcal{P}_a}{\text{min}} \;
        \left\lVert \bm{r} - \bm{r}^{\left( t \right)} \right\rVert_2,
\end{equation}
and computes the maximum error of the approximate Pareto frontier to the true Pareto frontier.
The MPFE measure can be sensitive to outliers, since a single bad point on Pareto frontier is sufficient to 
    negatively influence the measure.
VR is a measure that takes the entirety of the approximated Pareto frontier into consideration:
% EQUATION: VR
\begin{equation}
    \label{eq:vr}
    \text{VR} \left( \mathcal{P}_a \right) = -\log{
        \left( 1 - \frac{\text{Vol} \left( \mathcal{P}_a \right)}
        {\text{Vol} \left( \mathcal{P}_\text{true} \right)} \right)}
\end{equation}
with $\text{Vol}(\cdot)$ describing the hypervolume of the Pareto frontier.
The VR measure computes the ratio of hypervolume bounded by the approximate Pareto frontier vs. the 
    hypervolume bounded by the true Pareto frontier.
\review{We note that the VR measure is undefined if $\mathcal{P}_a$ exactly matches
    $\mathcal{P}_\text{true}$. However, since $\mathcal{P}_\text{true}$ itself is an approximation
    of the true Pareto frontier as stated above, this case is highly unlikely. The Equ.~\eqref{eq:vr}
    logarithm helps distinguish solutions with $\text{Vol} \left( \mathcal{P}_a \right)\ / \text{Vol}
    \left( \mathcal{P}_{\text{true}} \right) \simeq 1$, e.g.\ improvements of $\text{Vol}
    \left( \mathcal{P}_a \right)\ / \text{Vol} \left( \mathcal{P}_{\text{true}} \right)$ from 99 \% to 99.9 \%
    are valued higher than increasing the ratio from 19 \% to 19.9 \%.}

\subsection{Synthetic Benchmarks} \label{sec:synthetic_benchmarks}
\begin{figure*}
    \begin{center}
        \includegraphics[width=0.77\paperwidth]{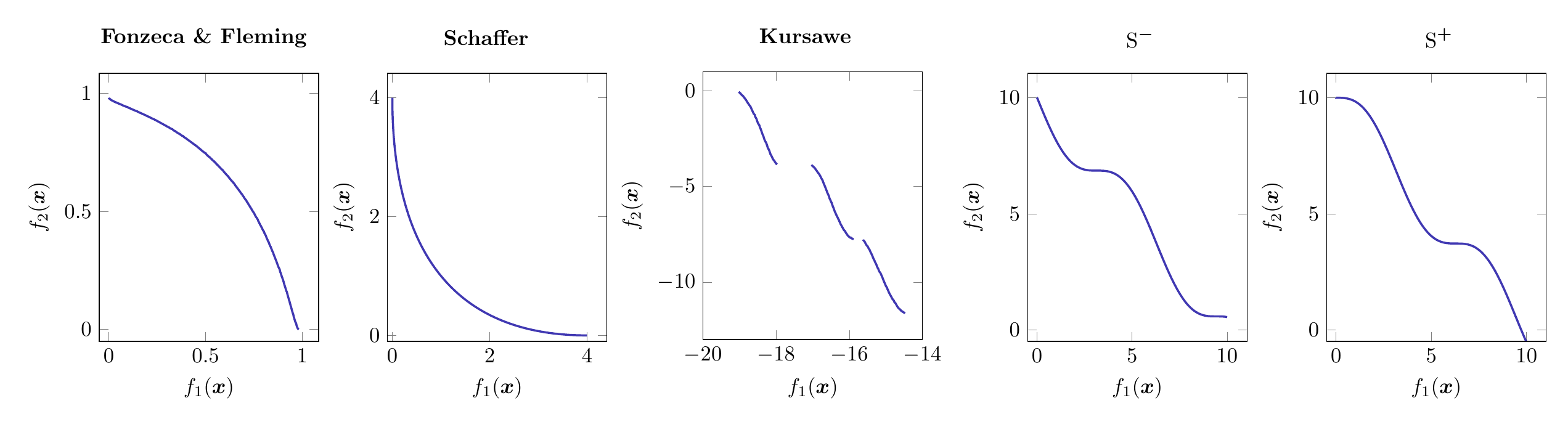}
    \end{center}
    \caption{\label{fig:syn_bench} Pareto frontiers of synthetic benchmark functions.}
\end{figure*}

We test the proposed method on various synthetic benchmarks to show its competitive performance 
    compared to other state-of-the-art methods.
The synthetic benchmarks provided in this section have different
    types of Pareto frontiers to show the algorithm's capability of reproducing them and each have two objective functions.
The \textit{Fonzeca \& Fleming} problem \citep{fonseca1995overview} is a popular choice for
    benchmarking multi-objective optimization frameworks and is defined as:
% EQUATION: fonzeca & fleming
\begin{equation}
    \label{eq:fonzeca}
    \begin{cases}
        f_1(\bm{x}) = 1 - \text{exp} \left( - \sum\limits_{i = 1}^{D}
            \left( x_i - \frac{1}{\sqrt {D}} \right)^2 \right), \\
        f_2(\bm{x}) = 1 - \text{exp} \left( - \sum\limits_{i = 1}^{D}
            \left( x_i + \frac{1}{\sqrt {D}} \right)^2 \right).
    \end{cases}
\end{equation}
\review{Fig.~\ref{fig:syn_bench}} shows the concave Pareto frontier of the function.
The function inputs $x_i \in \left[-4,4\right]$ for $i \in \{1,2,\dots,D\}$ bound the function outputs
    according to $f_1(\bm{x}), f_2(\bm{x}) \in \left[0,1\right]$.
For our tests we choose $D=2$. \par
The second synthetic benchmark function is \textit{Schaffer} \citep{schaffer1985some} given by:
% EQUATION: schaffer
\begin{equation}
    \label{eq:schaffer}
    \begin{cases}
        f_1(\bm{x}) = x^2, \\
        f_2(\bm{x}) = (x-2)^2.
    \end{cases}
\end{equation}
\textit{Schaffer} has a convex Pareto frontier as depicted in \review{Fig.~\ref{fig:syn_bench}}.
Inputs of \textit{Schaffer} are defined as $x \in \left[-3,3\right]$ bounding the objectives according to
    $f_1(x) \in \left[0,9\right]$ and $f_2(x) \in \left[0,25\right]$. \par
\textit{Kursawe} \citep{kursawe1990variant} is the third benchmark function which we use for comparison.
The \textit{Kursawe} function is interesting due to its discontinuous Pareto frontier and is defined as:
% EQUATION: kursawe
\begin{equation}
    \label{eq:kursawe}
    \begin{cases}
        f_1(\bm{x}) = \sum\limits_{i = 1}^{D-1}
            \left[ - 10 \text{exp} \left( -0.2 \sqrt {x_i^2 + x_{i+1}^2} \right)\right], \\
        f_2(\bm{x}) = \sum\limits_{i = 1}^{D}
            \left[ |x_i|^{0.8} + 5 \sin{\left( x_i^3 \right)} \right]
    \end{cases}
\end{equation}
\textit{Kursawe} has inputs defined as $x_i \in \left[-5,5\right]$ for $i \in \{1,2,\dots,D\}$ with $D=3$.
Approximate output bounds for \textit{Kursawe} are given as $f_1(\bm{x}) \in \left[-20,-4\right]$ and
    $f_2(\bm{x}) \in \left[-12,25\right]$. \par
We also include the $\text{S}^+$ and $\text{S}^-$  benchmark from \citet{olofsson2018bayesian} to test the 
    proposed algorithm on a problem that has both convex and concave sections on the Pareto frontier.
The functions are defined as follows:
% EQUATION: s+ / s-
\begin{equation}
    \label{eq:sin}
    \begin{cases}
        f_1(\bm{x}) = x_1, \\
        f_2(\bm{x}) = 10 - x_1 + x_2 \pm \sin\left( x_1 \right)
    \end{cases}
\end{equation}
where $\text{S}^+$ and $\text{S}^-$ are differentiated by the sign of the sinusoidal function in $f_2$. 
The two-dimensional inputs for both functions are given as $x_1 \in \left[0,10\right]$ and 
    $x_2 \in \left[0,10\right]$.
The approximate bounds of both objectives are $f_1(\bm{x}) \in \left[0,10\right]$ and 
    $f_2(\bm{x}) \in \left[-1,12\right]$. \par
Based on preliminary studies we found that the performance of different algorithms is highly dependent on 
    the quality of the initial population.
% rev2_16
\review{Therefore, we provide the same randomly sampled ten initial points to all
    methods for the synthetic benchmarks.}
% FIGURE: true pareto fronts
\review{Table~\ref{tab:syn_res}} presents the results of all synthetic benchmark tests.
% rev2_15
\review{The average of all random seeds for the proposed tree-based surrogate model outperforms
    \texttt{NSGA II} and random-search on all of the benchmark problems after 80 iterations for GD,
    IGD and VR metrics. For the MPFE metric the proposed method wins for every synthetic function
    except the Kursawe function.}
The results highlight the sampling-efficiency of surrogate-based methods.

% TABLE: results for synthetic benchmark problems
\pgfplotstableread[col sep=comma]{
    %FonzecaFleming.dat
    itr,entmoot_gd,entmoot_inv_gd,entmoot_mpfe,entmoot_vr,ga_gd,ga_inv_gd,ga_mpfe,ga_vr
    10,37.65 $\pm$ (7.7),4.15 $\pm$ (0.5),0.95 $\pm$ (0.1),0.03 $\pm$ (0.1),37.65 $\pm$ (7.7),4.15 $\pm$ (0.5),0.95 $\pm$ (0.1),0.03 $\pm$ (0.1)
    20,\textbf{16.90} $\pm$ (7.1),\textbf{2.93} $\pm$ (0.6),\textbf{0.67} $\pm$ (0.1),\textbf{0.42} $\pm$ (0.2),24.00 $\pm$ (7.8),3.59 $\pm$ (0.6),0.89 $\pm$ (0.1),0.10 $\pm$ (0.2)
    40,\:\:\textbf{4.09} $\pm$ (1.4),\textbf{1.51} $\pm$ (0.4),\textbf{0.45} $\pm$ (0.1),\textbf{1.47} $\pm$ (0.3),10.93 $\pm$ (7.7),2.83 $\pm$ (0.6),0.77 $\pm$ (0.2),0.51 $\pm$ (0.3)
    60,\:\:\textbf{2.20} $\pm$ (0.4),\textbf{1.10} $\pm$ (0.4),\textbf{0.38} $\pm$ (0.2),\textbf{2.10} $\pm$ (0.4),\:\:5.89 $\pm$ (6.4),2.01 $\pm$ (0.6),0.55 $\pm$ (0.1),0.78 $\pm$ (0.4)
    80,\:\:\textbf{1.48} $\pm$ (0.4),\textbf{0.87} $\pm$ (0.4),\textbf{0.30} $\pm$ (0.2),\textbf{2.54} $\pm$ (0.5),\:\:4.33 $\pm$ (4.8),1.55 $\pm$ (0.5),0.41 $\pm$ (0.1),1.29 $\pm$ (0.4)
    %Schaffer.dat
    10,25.66 $\pm$ (20.7),1.80 $\pm$ (0.5),1.56 $\pm$ (0.3),2.75 $\pm$ (0.7),25.66 $\pm$ (20.7),1.80 $\pm$ (0.5),1.56 $\pm$ (0.3),2.75 $\pm$ (0.7)
    20,\:\:\textbf{5.86} $\pm$ \:\:(3.3),\textbf{1.00} $\pm$ (0.1),\textbf{0.90} $\pm$ (0.2),\textbf{4.61} $\pm$ (0.4),\:\:7.54 $\pm$ \:\:(9.7),1.37 $\pm$ (0.4),1.11 $\pm$ (0.2),3.69 $\pm$ (0.6)
    40,\:\:\textbf{1.24} $\pm$ \:\:(0.9),\textbf{0.65} $\pm$ (0.1),\textbf{0.65} $\pm$ (0.2),\textbf{5.63} $\pm$ (0.5),\:\:1.58 $\pm$ \:\:(2.4),0.85 $\pm$ (0.3),0.81 $\pm$ (0.2),4.65 $\pm$ (0.6)
    60,\:\:\textbf{0.69} $\pm$ \:\:(0.3),\textbf{0.53} $\pm$ (0.1),\textbf{0.58} $\pm$ (0.2),\textbf{6.16} $\pm$ (0.6),\:\:0.92 $\pm$ \:\:(1.2),0.67 $\pm$ (0.2),0.64 $\pm$ (0.2),5.46 $\pm$ (0.5)
    80,\:\:\textbf{0.52} $\pm$ \:\:(0.2),\textbf{0.45} $\pm$ (0.1),\textbf{0.51} $\pm$ (0.2),\textbf{6.53} $\pm$ (0.7),\:\:0.59 $\pm$ \:\:(0.2),0.54 $\pm$ (0.1),0.56 $\pm$ (0.1),5.85 $\pm$ (0.4)
    %Kursawe.dat
    10,177.12 $\pm$ (32.6),21.15 $\pm$ (3.0),3.23 $\pm$ (0.3),0.55 $\pm$ (0.2),177.12 $\pm$ (32.6),21.15 $\pm$ (3.0),3.23 $\pm$ (0.3),0.55 $\pm$ (0.2)
    20,136.79 $\pm$ (25.1),20.08 $\pm$ (2.9),3.03 $\pm$ (0.3),\textbf{0.83} $\pm$ (0.2),\textbf{136.20} $\pm$ (23.2),\textbf{18.98} $\pm$ (2.6),\textbf{2.88} $\pm$ (0.3),0.73 $\pm$ (0.2)
    40,\:\: \textbf{61.70} $\pm$ (22.1),\textbf{13.84} $\pm$ (1.9),\textbf{2.47} $\pm$ (0.3),\textbf{1.41} $\pm$ (0.2),\:\:94.14 $\pm$ (18.4),17.30 $\pm$ (2.6),2.58 $\pm$ (0.3),0.99 $\pm$ (0.2)
    60,\:\: \textbf{32.17} $\pm$ (11.6),\textbf{12.37} $\pm$ (2.1),2.43 $\pm$ (0.3),\textbf{1.65} $\pm$ (0.2),\:\:78.84 $\pm$ (20.0),14.74 $\pm$ (2.8),\textbf{2.39} $\pm$ (0.3),1.23 $\pm$ (0.3)
    80,\:\: \textbf{23.62} $\pm$ \: (8.2),\textbf{10.71} $\pm$ (2.6),2.33 $\pm$ (0.4),\textbf{1.77} $\pm$ (0.3),\:\:57.72 $\pm$ (21.0),13.14 $\pm$ (2.4),\textbf{2.22} $\pm$ (0.3),1.41 $\pm$ (0.3)
    %Splus.dat
    10,27.41 $\pm$ (5.0),3.49 $\pm$ (0.4),1.77 $\pm$ (0.3),1.41 $\pm$ (0.2),27.41 $\pm$ (5.0),3.49 $\pm$ (0.4),1.77 $\pm$ (0.3),1.41 $\pm$ (0.2)
    20,\textbf{17.18} $\pm$ (2.8),\textbf{2.77} $\pm$ (0.2),1.41 $\pm$ (0.1),1.90 $\pm$ (0.2),19.14 $\pm$ (3.6),2.85 $\pm$ (0.4),\textbf{1.40} $\pm$ (0.3),\textbf{1.92} $\pm$ (0.2)
    40,\textbf{10.12} $\pm$ (1.0),\textbf{2.10} $\pm$ (0.2),\textbf{1.08} $\pm$ (0.2),2.26 $\pm$ (0.3),13.73 $\pm$ (2.7),2.24 $\pm$ (0.3),1.21 $\pm$ (0.2),\textbf{2.29} $\pm$ (0.2)
    60,\:\:\textbf{7.27} $\pm$ (0.9),\textbf{1.76} $\pm$ (0.1),\textbf{0.96} $\pm$ (0.1),\textbf{2.60} $\pm$ (0.4),10.69 $\pm$ (2.6),2.08 $\pm$ (0.2),1.06 $\pm$ (0.2),2.47 $\pm$ (0.2)
    80,\:\:\textbf{5.24} $\pm$ (0.5),\textbf{1.57} $\pm$ (0.1),\textbf{0.88} $\pm$ (0.1),\textbf{2.78} $\pm$ (0.4),\:\:8.37 $\pm$ (2.5),1.92 $\pm$ (0.1),1.01 $\pm$ (0.1),2.73 $\pm$ (0.2)
    %Sminus.dat}
    10,27.93 $\pm$ (3.8),3.26 $\pm$ (0.2),1.65 $\pm$ (0.2),1.35 $\pm$ (0.2),27.93 $\pm$ (3.8),3.26 $\pm$ (0.2),1.65 $\pm$ (0.2),1.35 $\pm$ (0.2)
    20,\textbf{18.12} $\pm$ (1.9),\textbf{2.61} $\pm$ (0.2),1.31 $\pm$ (0.1),\textbf{1.92} $\pm$ (0.1),19.56 $\pm$ (2.6),2.68 $\pm$ (0.2),\textbf{1.30} $\pm$ (0.2),1.81 $\pm$ (0.1)
    40,\textbf{10.35} $\pm$ (1.2),\textbf{2.02} $\pm$ (0.1),\textbf{1.10} $\pm$ (0.1),\textbf{2.55} $\pm$ (0.1),13.73 $\pm$ (2.3),2.29 $\pm$ (0.2),1.14 $\pm$ (0.1),2.15 $\pm$ (0.1)
    60,\:\:\textbf{7.40} $\pm$ (0.7),\textbf{1.68} $\pm$ (0.1),\textbf{0.97} $\pm$ (0.1),\textbf{2.93} $\pm$ (0.1),10.39 $\pm$ (2.5),2.08 $\pm$ (0.1),1.05 $\pm$ (0.1),2.35 $\pm$ (0.1)
    80,\:\:\textbf{5.38} $\pm$ (0.5),\textbf{1.51} $\pm$ (0.1),\textbf{0.89} $\pm$ (0.1),\textbf{3.21} $\pm$ (0.1),\:\:8.17 $\pm$ (2.3),1.95 $\pm$ (0.1),1.00 $\pm$ (0.1),2.49 $\pm$ (0.1)
}\synRes

\begin{table}[]
    \pgfplotstableset{col sep=comma}

    \pgfplotstableset{create on use/Problem/.style={
            create col/set list={
                ,Fonzeca \&, Fleming \citep{fonseca1995overview},,,
                ,Schaffer \citep{schaffer1985some},,,,
                ,Kursawe \citep{kursawe1990variant},,,,
                ,S+ \citep{olofsson2018bayesian},,,,
                ,S- \citep{olofsson2018bayesian},,,,
                },
        },
        columns/Problem/.style={string type, column type={|c}},
    }
    
    \pgfplotstabletypeset[
        precision=2,
        fixed zerofill, column type={r},
        string type,
        every head row/.style={%
            before row={\hline
                & & \multicolumn{2}{c|}{GD ($\times 100$)}
                & \multicolumn{2}{c|}{inv. GD ($\times 100$)}
                & \multicolumn{2}{c|}{MPFE} & \multicolumn{2}{c|}{VR}\\
            },
            after row=\hline
        },
        font=\scriptsize,
        every last row/.style={after row=\hline},
        columns={Problem, itr, entmoot_gd, ga_gd, entmoot_inv_gd, ga_inv_gd,
            entmoot_mpfe, ga_mpfe, entmoot_vr, ga_vr},
        % itr
        columns/itr/.style={column name=Itr, column type={c}},
        % gd
        columns/entmoot_gd/.style={column name=\texttt{ENTMOOT}, column type={|c}},
        columns/ga_gd/.style={column name=\texttt{NSGA2} / rnd, column type={c}},
        % inv. gd
        columns/entmoot_inv_gd/.style={column name=\texttt{ENTMOOT}, column type={|c}},
        columns/ga_inv_gd/.style={column name=\texttt{NSGA2} / rnd, column type={c}},
        % mpfe
        columns/entmoot_mpfe/.style={column name=\texttt{ENTMOOT}, column type={|c}},
        columns/ga_mpfe/.style={column name=\texttt{NSGA2} / rnd, column type={c}},
        % vr
        columns/entmoot_vr/.style={column name=\texttt{ENTMOOT}, column type={|c}},
        columns/ga_vr/.style={column name=\texttt{NSGA2} / rnd, column type={c|}},
        % seperate tables
        every row no 5/.style={before row=\hline},
        every row no 10/.style={before row=\hline},
        every row no 15/.style={before row=\hline},
        every row no 20/.style={before row=\hline},
        ]{\synRes}
    \caption{
    \label{tab:syn_res}
    Results of synthetic benchmark problems comparing \texttt{ENTMOOT} against the best result
        of \texttt{NSGA-II} and random-search based on different performance metrics. 
    Each entry shows the Median of all runs with different random seeds and the standard deviation in brackets after a certain 
        amount of black-box evaluations with winning median highlighted in bold.}
\end{table}

\subsection{Optimization of Windfarm Layout} \label{sec:wind}
\begin{table}[]
    \centering
    \begin{tabularx}{\textwidth}{ c|X }
        \hline
        $u_j$ & experienced velocity of turbine $j$ \\
        $u_0$ & ambient wind speed \\
        $u_{k,j}$ & interference caused on turbine $j$ by turbine $k$ \\
        $C_{T,k}$ & thrust coefficient of $k$-th turbine at given windspeed \\
        $A_j$ & rotor area of $j$-th turbine \\
        $A_{k,j}$ & rotor area of $j$-th turbine lying in the wake of turbine $k$ \\
        $h_\text{hub}$ & turbine hub height \\
        $z_0$ & surface roughness height \\
        $\text{dist}_{k,j}$ & distance between turbines $k$ and $j$ \\
        $\xi$ & degree of wake expansion \\
        $R_k$ & radius of wind turbine $k$ \\
        $c_{k,j}$ & distance from turbine $j$ to center of wake from turbine $k$ when it 
             reaches turbine $j$ \\
        $f_w$ & annual frequency at which certain wind speeds occur \\
        $P_j(\cdot)$ & power generated by turbine $j$ \\
        $P_j^\text{ideal}(\cdot)$ & power generated by turbine $j$ without wake effects\\
        $x_\text{turb},k$ & x-coordinates of turbine $k$ \\
        $y_\text{turb},k$ & y-coordinates of turbine $k$ \\
        $b_k$ & binary variable indicating if turbine $k$ is active \\
        $b_{k,j}^\text{dist}$ & auxiliary binary variable indicating if distance constrained for 
            turbines $k$ and $j$ is active \\
        $\text{dist}_{k,j}$ & continuous variable for distance between turbines $k$ and $j$, 
            change with definition above \\
        $\alpha_\text{samp}$ & continuous variable that gives
            maximum distance to closest sampling point \\
        $k_i^{d,+}$ & auxiliary variable takes positive contribution of Manhattan distance 
            between $x_i$ and data point $d$ \\
        $k_i^{d,-}$ & auxiliary variable takes negative contribution of Manhattan distance 
            between $x_i$ and data point $d$ \\
        $N_\text{turb}$ & number of active turbines \\
        \hline
    \end{tabularx}
    \caption{Wind turbine benchmark table of notation.}
    \label{tab:wind_turbine_notation}
\end{table}

\subsubsection{General Model Description}
In this section, we consider the optimal layout of an offshore windfarm based on the model by 
    \citet{rodrigues2016multi}. 
In general, formulations for optimization of windfarm layouts can be nonlinear, multi-modal, 
    non-differentiable, non-convex, and discontinuous, making them difficult to solve using model-based 
    optimization techniques. 
Therefore, the resulting problems are often solved using multi-objective black-box optimization schemes. 
We note that, while the model presented here is based on several simplifications, the proposed optimization strategy is 
    general and can be applied to more complicated windfarm models, including those that cannot be 
    expressed analytically. 

The placement of wind turbines is complicated by wake effects, or the effect of one turbine on the power
    generation of another. 
Wake losses are approximated using the model by \citet{katic1986simple}, which provides a simple model for 
    approximating wind velocity in the wake of some turbine(s).
Table~\ref{tab:wind_turbine_notation} provides the notation for this case study.

Following this model, the velocity experienced by the $j$-th turbine is expressed:
\begin{equation}
    u_j = u_0(1-\sqrt{\sum_{k \neq j} u_{k,j}^2})
\end{equation}
where $u_0$ is the ambient wind speed (without wake effects) and $u_{k,j}$ is the interference caused on 
    turbine $j$ by turbine $k$. 
The pairwise interference terms are calculated as:
\begin{align}
    u_{k,j} &= \frac{1-\sqrt{1-C_{T,k}}}{(1+\xi \text{dist}_{k,j}/R_j)^2} \frac{A_{k,j}}{A_j} \\
    \xi &= 0.5 \mathrm{log}^{-1}\left( \frac{h_\mathrm{hub}}{z_0} \right)
\end{align}
where $C_{T,k}$ is the thrust coefficient of the $k$-th turbine at a given windspeed, $A_j$ is the rotor 
    area of the $j$-th turbine, and $A_{k,j}$ is the rotor area of the $j$-th turbine that lies in the wake
    of turbine $k$. 
The variable $\text{dist}_{k,j}$ denotes distance between turbines $k$ and $j$, and $\xi$ represents the 
    degree of wake expansion, computed using turbine hub height $h_\mathrm{hub} = 107$ m, and 
    surface roughness height $z_0$, assumed to be $5\times10^{-4}$ m. 
It is further assumed that the wake of each turbine expands linearly, i.e., $R_{k,j} = R_k + \xi
    \text{dist}_{k,j}$, where $R_k$ is the radius of the turbine rotor. 
Therefore, the rotor area of turbine $j$ in the wake of turbine $k$, $A_{k,j}$, can be computed as:
\begin{multline}
    A_{k,j} = \frac{1}{2} \left( R_{k,j}^2 \left( 2 \mathrm{arccos}\left( \frac{R_{k,j}^2 + c_{k,j}^2 - 
    R_j^2}{2 R_{k,j} c_{k,j}} \right) - \mathrm{sin} \left( 2 \mathrm{arccos}\left( \frac{R_{k,j}^2 + 
    c_{k,j}^2 - R_j^2}{2 R_{k,j} c_{k,j}} \right) \right) \right) \right) + \\
    \frac{1}{2} \left( R_{j}^2 \left( 2 \mathrm{arccos}\left( \frac{R_{j}^2 + c_{j,k}^2 - R_{k,j}^2}{2 
    R_{j} c_{j,k}} \right) - \mathrm{sin} \left( 2 \mathrm{arccos}\left( \frac{R_{j}^2 + c_{k,j}^2 - 
    R_{k,j}^2}{2 R_{j} c_{j,k}} \right) \right) \right) \right)
\end{multline}
where $c_{k,j}$ is the distance from turbine $j$ to the center of the wake from turbine $k$ when it reaches
    turbine $j$. 
Note that if the wake front entirely covers turbine $j$, then instead $A_{k,j}$ is set to $\pi R_j^2$, and
    if the wake front does not impact turbine $j$, then $A_{k,j} = 0$. 

Wind turbines are assumed to be identical ``Vestas 8 MW'' units, whose power and thrust curves are linearly
    interpolated and given in \citet{rodrigues2016multi}. Each turbine has a rotor radius of 82 m. 
We further use the discretized North Sea wind distribution data from \citet{rodrigues2016multi}.
% rev_1_3, rev_1_4, rev_1_5
\review{This application places the base of up to 16 wind turbines within a square 15.21~$\text{km}^2$
    area based on optimal tradeoffs of two conflicting objectives to explore the Pareto frontier.}
We represent the coordinates of each turbine using two continuous variables; therefore, the degrees of
    freedom for the \textit{windfarm} problem $\bm{x}$ comprise 32 positive continuous variables, i.e.\ 
    Equ.~\eqref{eq:req_constr_f} and Equ.~\eqref{eq:req_constr_g}, denoting the location of each of the
    16 wind turbines.
Furthermore, 16 binary variables, i.e.\ Equ.~\eqref{eq:const_e}, are introduced to indicate which wind 
    turbines are active.

Optimization of the windfarm layout involves two objective functions:
\begin{equation}
    \begin{cases}
        f_1(\bm{x}) = \frac{\sum_w \sum_{j=1}^{16} P_j(u_w) f_w}{\sum_w 16 P_j^\mathrm{ideal}(u_w) f_w}, \\
        f_2(\bm{x}) = \frac{\sum_w \sum_{j=1}^{16} P_j(u_w) f_w}{N_\text{turb} P_j^\mathrm{ideal}}.
    \end{cases}
\end{equation}
where $P_j(u_w)$ is the power generated by turbine $j$ generated at wind velocity $u_w$ (wind data is 
    discretized in to winds with velocity $u_w$ and frequency $f_w$). 
$P_j^\mathrm{ideal}(u_w)$ is the power generated by turbine $j$ at wind velocity $u_w$ without any wake 
    effects, and $N_\text{turb}$ is the number of turbines placed in the farm. 
The objectives $f_1$ and $f_2$ correspond to the energy production and efficiency, respectively. 
The two objectives represent conflicting design goals, as production is increased by placing more turbines 
    in the wind farm. 
However, wake effects also increase with the number of turbines, decreasing the windfarm efficiency. 

\subsubsection{Optimization Benchmark} \label{sec:wind_opt_bench}
% FIGURE: true pareto fronts
\begin{figure*}
    \begin{center}
        \includegraphics[width=0.55\paperwidth]
            {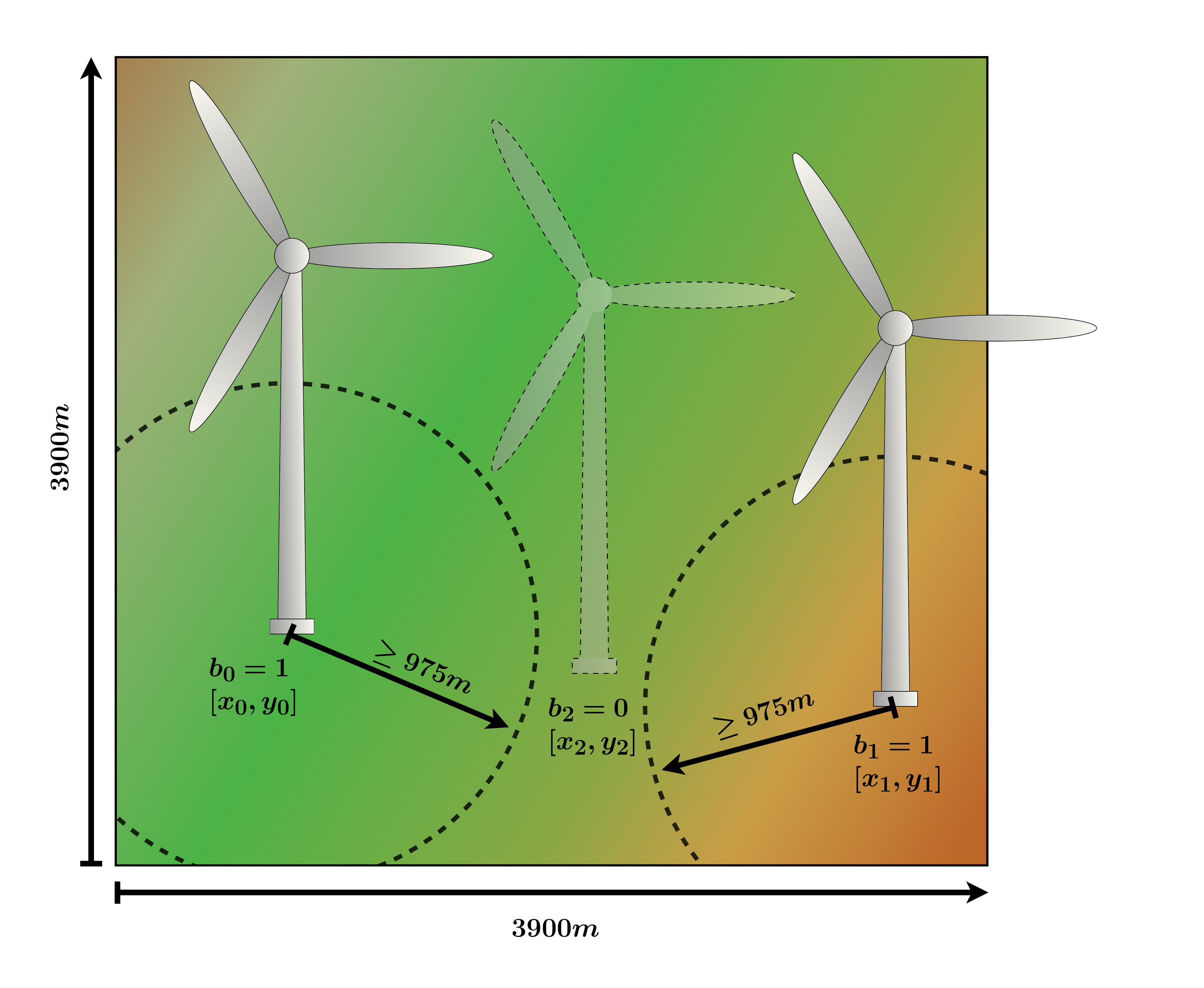}
    \end{center}
    \caption{\label{fig:ex_wind} \review{Example for feasible layout using three wind turbines
        visualizing optimization variables from Equ.~\eqref{eq:req_constr}. Variabels $b$ indicate if
        a wind turbine is active and $[x,y]$ give its coordinates.
    Circles around wind turbines indicate the Equ.~\eqref{eq:req_constr_b} minimum distance constraints
        of 975 m between individual wind turbines.}}
\end{figure*}
For the \textit{windfarm} benchmark we distinguish between required and supporting constraints, i.e.\ 
    Equ.~\eqref{eq:req_constr} and Equ.~\eqref{eq:help_constr}, respectively.
Required constraints are given by the problem description and must be satisfied by feasible solutions.
Equ.~\eqref{eq:req_constr_b} define auxiliary variables $\text{dist}_{i,j}$ as the distance between two 
    wind turbine locations and Equ.~\eqref{eq:req_constr_c} defines the minimum of distance 
    $\text{dist}_{k,j}$ as 975 m when both turbines $k$ and $j$ are active which is given when 
    Equ.~\eqref{eq:req_constr_d} is zero.
\review{Fig.~\ref{fig:ex_wind}} depicts a feasible layout for three wind turbines.
Moreover, supporting constraints in Equ.~\eqref{eq:help_constr} are added to break symmetry and
    remove equivalent solutions.
Equ.~\eqref{eq:help_constr_a} assures that wind turbines are activated in an increasing order, e.g.\ for
    the case that two wind turbines are active this allows only for binary variables $b_1$ and $b_2$ to 
    be active.
The inequality constraints Equ.~\eqref{eq:help_constr_b} and Equ.~\eqref{eq:help_constr_c} enforce 
    a default location of $(0,0)$ for inactive wind turbines since their location has no influence 
    on the black-box objective.
We found that \texttt{NSGA-II} struggles to consistently provide feasible solutions
    for the required constraints in Equ.~\eqref{eq:req_constr}, so we exclude supporting constraints from 
    Equ.~\eqref{eq:help_constr} because they further restrict the search space.
Since \texttt{ENTMOOT} and the initial sampling set both provide guaranteed $\epsilon$-feasible solutions,
    additional supporting constraints from Equ.~\eqref{eq:help_constr} can be added to further limit the 
    search space and prevent the algorithm from finding redundant solutions. \\
% added constraints
\textbf{Required Constraints}
\begin{subequations}
    \label{eq:req_constr}
    \begin{flalign}
        & \sum\limits_{k = 1}^{|\bm{b}|} b_k \geq 1, & & \label{eq:req_constr_a}\\
        & \text{dist}_{k,j}^2 \leq
            \left( x_{\text{turb},k} - x_{\text{turb},j} \right)^2 +
            \left( y_{\text{turb},k} - y_{\text{turb},j} \right)^2,
            & & \forall k \in \left[ 1, ..., 16 \right], \;
            \forall j \in \left[ k + 1, ..., 16 \right] \label{eq:req_constr_b}\\
        & b_{k,j}^{\text{dist}} \rightarrow \text{dist}_{i,j} \geq 975,
            & & \forall k \in \left[ 1, ..., 16 \right], \;
            \forall j \in \left[ i + 1, ..., 16 \right], \label{eq:req_constr_c} \\
        & b_{k,j}^{\text{dist}} = b_k \cdot b_j,
            & & \forall k \in \left[ 1, ..., 16 \right], \;
            \forall j \in \left[ k + 1, ..., 16 \right], \label{eq:req_constr_d} \\
        & \bm{b} \in \left\{ 0, 1 \right\}^{16}, & & \label{eq:req_constr_e} \\
        & \bm{x}_{\text{turb}} \in \left[ 0.0, 3900.0 \right]^{16}, & & \label{eq:req_constr_f} \\
        & \bm{y}_{\text{turb}} \in \left[ 0.0, 3900.0 \right]^{16} & & \label{eq:req_constr_g} \\
        & \text{dist}_{k,j} \in \mathbb{R}^\text{+},
            & & \forall k \in \left[ 1, ..., 16 \right], \;
            \forall j \in \left[ k + 1, ..., 16 \right] \label{eq:req_constr_h}\\
        & b_{k,j}^{\text{dist}} \in \left\{ 0, 1 \right\},
            & & \forall k \in \left[ 1, ..., 16 \right], \;
            \forall j \in \left[ k + 1, ..., 16 \right]. \label{eq:req_constr_i}
    \end{flalign}
\end{subequations}

\textbf{Helper Constraints}
\begin{subequations}
    \label{eq:help_constr}
    \begin{flalign}
        & b_k \geq b_{k+1},
            & & \forall k \in \left[ 1, ..., 15 \right], \label{eq:help_constr_a} \\
        & \neg b_i \rightarrow x_{\text{turb},k} \leq 0,
            & & \forall k \in \left[ 1, ..., 16 \right], \label{eq:help_constr_b} \\
        & \neg b_i \rightarrow y_{\text{turb},k} \leq 0,
            & & \forall k \in \left[ 1, ..., 16 \right]. \label{eq:help_constr_c}
    \end{flalign}
\end{subequations}
We use an advanced initialisation strategy to ensure that only feasible initial populations are provided 
    for both \texttt{ENTMOOT} and \texttt{NSGA-II}.
Moreover, the solutions are created for fixed numbers of wind turbines.
The goal is to provide 16 initial solutions with $1, 2, ..., 16$ wind turbines to expose both 
    tested algorithms to initial points that contain all possible numbers of 
    wind turbines.
The algorithm to compute the initial set of data points $\mathcal{D}_\text{init}$ is given as
    Algorithm~\ref{algo:init}.
The first data point has one turbine placed at the middle of the field with random 
    noise $\epsilon_\text{noise}$ added to the its position to get varying results for different random 
    seeds.
Equ.~\eqref{eq:init} takes into consideration all data points that are currently in
    $\mathcal{D}_\text{init}$ and maximizes the Manhattan distance to the closest sample 
    that is available.
We encode the Manhattan distance according to \citet{giloni2002L1}. 
Equ.~\eqref{eq:man_c} enforces $k_i^{d,+}$ and $k_i^{d,-}$ to take positive and negative values of
    the difference between $x_i$ and the closest data point, respectively.
Equ.~\eqref{eq:const_d} is implemented using special ordered sets \citep{tomlin1988SOS} to assure that at 
    least one of the variables $k_i^{d,+}$ and $k_i^{d,-}$ is zero.
The fixed number $N_\text{turb}$ of turbines per iteration is enforced through the summation
    constraint Equ.~\eqref{eq:man_f}. \\
    
\begin{algorithm}[H]
    \begin{algorithmic}[1]
        \Require{$seed \in \mathbb{Z}^+$}
        \Ensure{$\mathcal{D}_\text{init}$}
        \State{$\mathcal{D}_\text{init} \gets \{\}$}
        \State{$N_\text{turb} \gets 1$}
        \State{$\epsilon_\text{noise} \gets$ random$\left(0, 1950.0, seed\right)$}
        \State{$x_{\text{turb},0} \gets 1950.0 + \epsilon_\text{noise}$}
        \State{$y_{\text{turb},0} \gets 1950.0 + \epsilon_\text{noise}$}
        \State{$\bm{x}_{\text{init}} \gets$ Solve Equ.~\eqref{eq:init} s.t.\
            Equ.~\eqref{eq:req_constr} and Equ.~\eqref{eq:help_constr} with 
            $x_{\text{turb},0}$,
            $y_{\text{turb},0}$ and
            $N_\text{turb}$}
        \State{$\mathcal{D}_\text{init} \gets \mathcal{D}_\text{init} \bigcup \bm{x}_{\text{init}}$}
        \For{$l \gets 2$ to $16$}
            \State{$N_\text{turb} \gets l$}
            \State{$\bm{x}_{\text{init}} \gets$ Solve Equ.~\eqref{eq:init} s.t.\
                Equ.~\eqref{eq:req_constr} and Equ.~\eqref{eq:help_constr}
                with $\mathcal{D}_\text{init}$ and $N_\text{turb}$}
            \State{$\mathcal{D}_\text{init} \gets \mathcal{D}_\text{init} \bigcup \bm{x}_{\text{init}}$}
        \EndFor
    \end{algorithmic}
     \caption{Wind turbine initialisation algorithm.}
     \label{algo:init}
\end{algorithm}
% EQUATIONS: manhattan distance
\begin{subequations}
    \label{eq:init}
    \begin{flalign}
        & & \bm{x}_{\text{init}} \in \underset{\bm{x}} {\text{arg min}} & \;
            - \alpha_\text{samp} & \\
        & & \text{s.t.} \; \; & \alpha_\text{samp} \leq 
            \sum\limits_{i \in \left [ n \right ]}
            k_i^{d,+} + k_i^{d,-},
            & & \forall d\in \mathcal{D}_\text{init}, \label{eq:man_b}\\
        & & & x_i^{d,\text{norm}} - \frac{x_i - v^L_i}{v^U_i - v^L_i} = 
            k_i^{d,+} - k_i^{d,-},
            & & \forall d\in \mathcal{D}_\text{init},
            \forall i \in \left [ n \right ],\label{eq:man_c}\\
        & & & k_i^{d,+} \cdot k_i^{d,-} = 0,
            & & \forall d\in \mathcal{D}_\text{init},
            \forall i \in \left [ n \right ], \label{eq:man_d}\\
        & & & k_i^{d,+}, k_i^{d,-} \geq 0,
            & & \forall d\in \mathcal{D}_\text{init},
            \forall i \in \left [ n \right ], \label{eq:man_e}\\
        & & & \sum\limits_{i = 1}^{|\bm{b}|} b_i = N_\text{turb}. & & \label{eq:man_f}
    \end{flalign}
\end{subequations}
% FIGURE: windfarm results
\begin{figure*}
    \begin{center}
        \includegraphics[width=0.78\paperwidth]{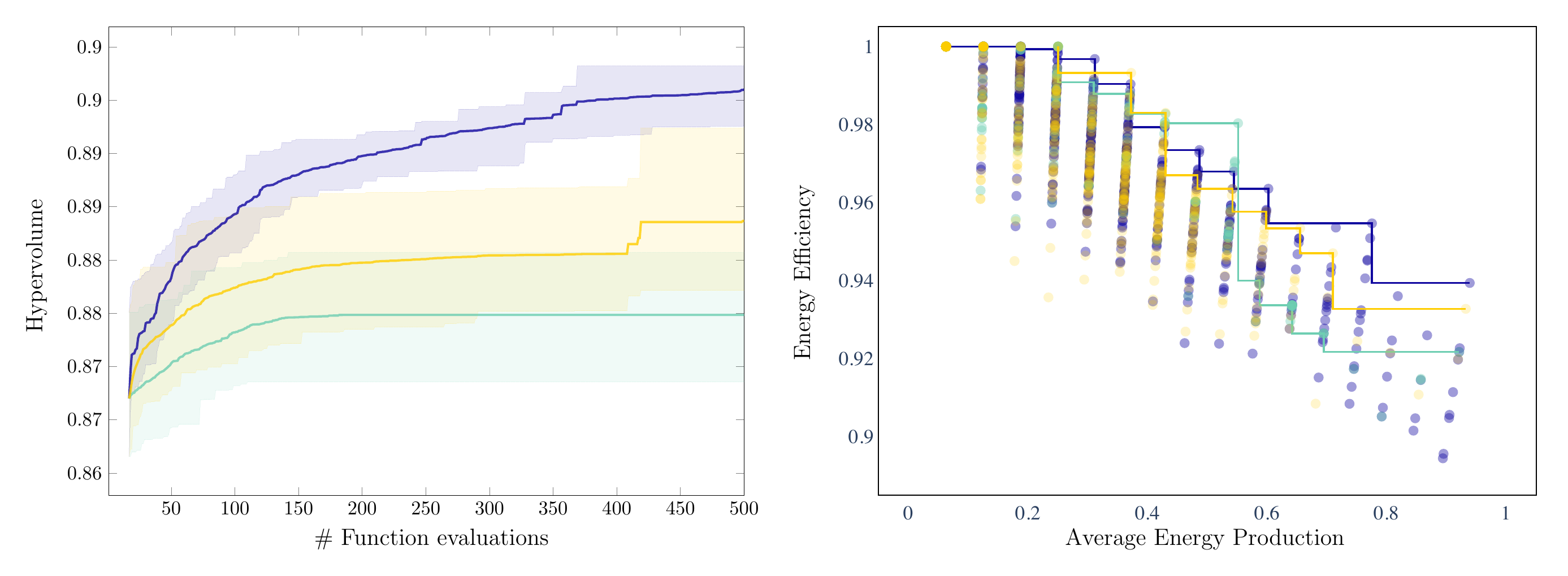}
    \end{center}
    \caption{Compares windpark layout optimization results of \texttt{NSGA-II} (green),
    \texttt{ENTMOOT} (purple) and feasible sampling strategy (yellow). \textbf{(left)}~Hypervolume
    bounded by Pareto frontier approximation after a certain number of function evaluations. 
    \textbf{(right)}~Best Pareto frontier approximation found for different algorithms based on 
    hypervolume. Objectives are dimensionless due normalization based on physical limits.}
    \label{fig:res_windfarm}
\end{figure*}

% FIGURE: windfarm layout proposals for
\begin{figure*}
    \begin{center}
        \includegraphics[width=0.80\paperwidth]
            {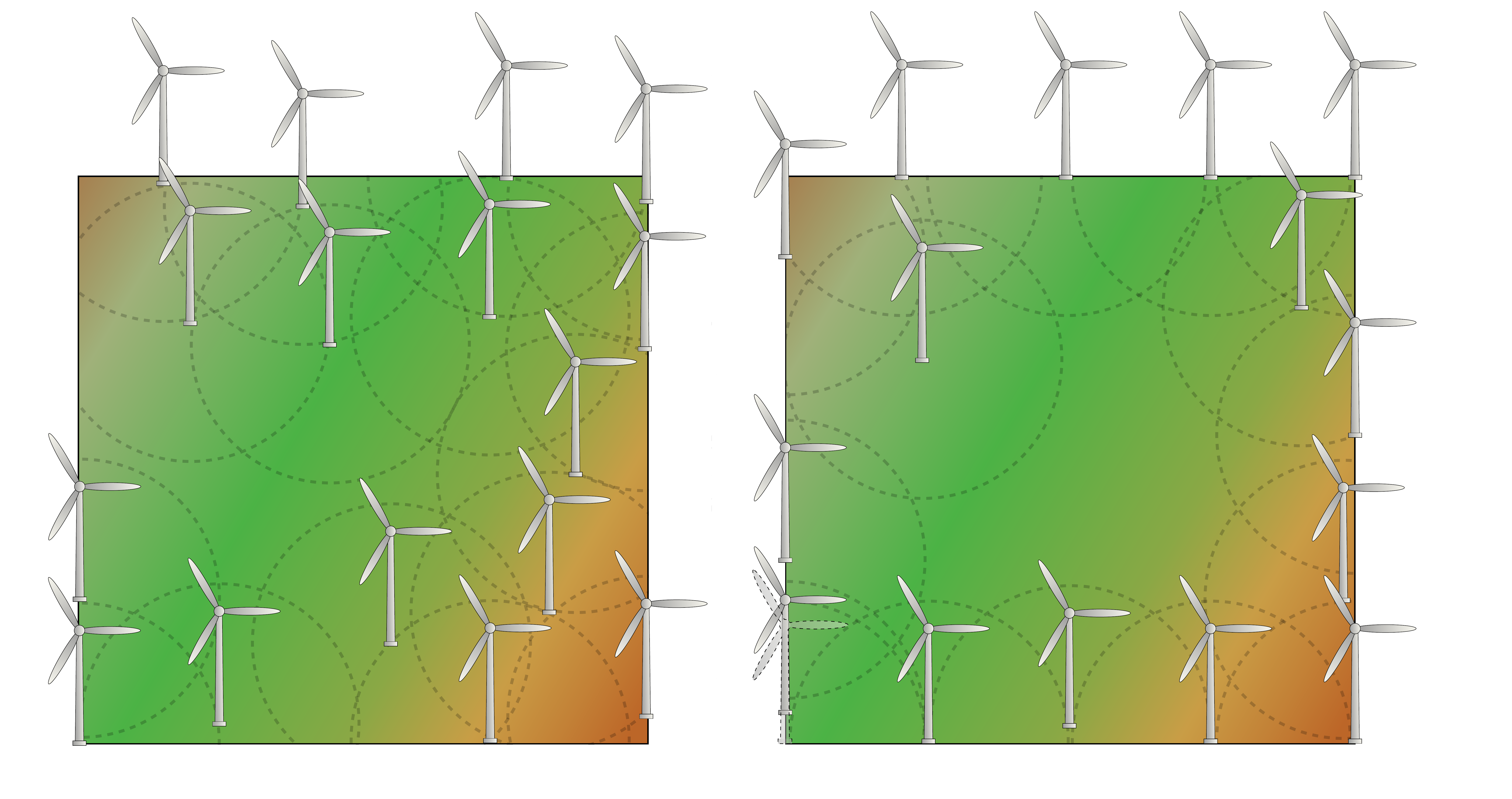}
    \end{center}
    \caption{\review{\label{fig:res_wind_layout} Example windpark layouts for highest average energy objective
    solutions not including initial samples, i.e. \textbf{(left)} \texttt{ENTMOOT} and
    \textbf{(right)} \texttt{NSGA-II}.
    Circles around wind turbines indicate the Equ.~\eqref{eq:req_constr_b} minimum distance constraints
        of 975 m between individual wind turbines. \texttt{ENTMOOT} is able to place all 16 turbines while \texttt{NSGA-II} is only able to place 15 due to constraint violation.}}
\end{figure*}

Fig.~\ref{fig:res_windfarm} shows the bounded hypervolume for different algorithms for a total of 500 
    black-box evaluations.
For better comparison we include test runs for the initial sampling strategy where 500 total samples are 
    generated.
\texttt{NSGA-II} finds significantly better results early on when compared to a similar study presented in
    \citep{rodrigues2016multi}, which is likely due to variety of initial feasible points provided, 
    since GAs generate new black-box input proposals based on the previously observed 
    population.
However, even the feasible sampling strategy quickly outperforms \texttt{NSGA-II}, emphasizing
    the importance of providing feasible black-box input proposals.
This is even more evident for solutions with large average energy production and low energy efficiency since these refer to windfarm
    layouts with a high number of wind turbines where more energy is produced with low energy efficiency due to wake effects.
From an optimization perspective such solutions are more difficult due to Equ.~\eqref{eq:req_constr_b} minimum distance 
    constraints enforced between all turbines that become increasingly restrictive for a growing number of wind turbines.
\review{Fig.~\ref{fig:res_wind_layout} depicts layout proposals given by \texttt{ENTMOOT} and \texttt{NSGA-II}
    for highest average energy production, i.e.\ solutions with high number of turbines,
    when excluding the set of initial samples.
While \texttt{NSGA-II} fails to place all 16 turbines due to constraint violation, \texttt{ENTMOOT}
    manages to find a feasible solution for all turbines available highlighting its capability to
    explore the entire Pareto frontier of non-dominated objective trade-offs.
\review{Moreover, \texttt{ENTMOOT} profits from its capability to explicitly handle input constraints.
The presented 32-dimensional case study has a smaller effective dimensionality since only positions
    of active turbines influence the black-box outcome, e.g. if only one turbine is active,
    the problem becomes three dimensional.
Since \texttt{ENTMOOT} can capture this hierarchical structure,
    it never revisits equivalent solutions and better uses the sampling budget.}
Fig.~\ref{fig:res_windfarm} depicts the best Pareto frontier approximations found by each algorithm and shows \texttt{ENTMOOT}'s 
    outstanding performance compared to \texttt{NSGA-II} for high average energy production solutions due to guaranteed feasible 
    input proposals even for highly-constrained problems.}
\texttt{ENTMOOT} also outperforms the feasible sampling strategy, highlighting the 
    significance of using tree ensembles as surrogate models to learn multi-objective systems.
Note that \citet{rodrigues2016multi} only tested on a discrete grid with varying resolution because 
    continuous coordinates were deemed too difficult.
With 500 iterations, \texttt{ENTMOOT} for the continuous case outperforms all methods presented in \citep{rodrigues2016multi} 
    for the highest discretization resolution even after up to $10^6$ iterations.
While one can argue that the windfarm layout benchmark presented here is cheap to evaluate, 
    real-world layout planning will require many expensive dynamic fluid simulations, and sampling-efficient
    algorithms can help with quickly finding good solutions

\subsection{Battery Material Optimization}

\begin{table}[]
    \centering
    \begin{tabularx}{\textwidth}{ c|X }
        \hline
        $p$ & parameter set picked for default values \\
        $\epsilon_\text{poros}^+$ & porosity of positive battery side \\
        $\epsilon_\text{active}^+$ & active material volume fraction of positive battery side \\
        $R_\text{particle}^+$ & particle radius of positive battery side \\
        \review{$C$} & \review{current that discharges the battery in one hour} \\
        $\lambda^+$ & electrode thickness scaling factor of positive battery side \\
        $\epsilon_\text{poros}^-$ & porosity of negative battery side \\
        $\epsilon_\text{active}^-$ & active material volume fraction of negative battery side \\
        $R_\text{particle}^-$ & particle radius of negative battery side \\
        $\lambda^-$ & electrode thickness scaling factor of negative battery side \\
        \review{$I$} & \review{discharge current} \\
        \review{$V$} & \review{discharge voltage} \\
        \review{$t$} & \review{discharge time} \\
        \hline
    \end{tabularx}
    \caption{Battery benchmark table of notation.}
    \label{tab:battery_notation}
\end{table}

\subsection{General Model Description}
Battery technology is a popular field of research and there are many examples in literature that 
    successfully model key performance metrics, e.g.\ capacity degradation \citep{severson2019data}, 
    by using data-driven approaches.
In this section we consider Lithium-ion battery optimization: namely how to simultaneously achieve high energy and power density. For this exercise the open-source software PyBaMM \citep{Sulzer2021} is used, which comprises a number of mathematical models for simulating the important physical processes that occur within batteries. PyBaMM makes extensive use of the CasADi framework \citep{Andersson2019} to solve the system of DAEs and the NumPy \citep{Harris2020} python package for array manipulation. Within the PyBaMM framework, the Single Particle Model with electrolyte effects (SPMe) \citep{Marquis2019} is chosen as it offers a good compromise between computational cost and accuracy (including electrolyte losses and particle diffusion). 
While recent works \citep{Tranter2020a} and \citep{Timms2021} have studied battery-state heterogeneity, we select a one-dimensional, isothermal formulation here for simplicity, noting that the proposed framework is agnostic to the form of the black-box model. 
The full suite of battery models available in PyBaMM is described in \citet{Marquis2020}.

A typical Lithium-ion battery comprises layers of electrodes containing active material that host the ions at different open-circuit potentials. 
These electrodes are separated by non-conductive, electrolyte filled polymers, with layers alternating between anode (negative) and cathode (positive). 
The electrodes are usually coated onto both sides of metallic foils that form the current collectors, which are typically copper and aluminium. The battery charges and discharges when an external current is passed between the cell terminals, and ions flow internally within the electrolyte between the electrodes, completing redox reactions at the active-material surfaces. Each electrode is formed of a solid matrix mixture of active material, to support the intercalation of lithium ions, and carbon binder which structurally supports the active material and conducts electrons towards the reaction sites. The matrix is porous, allowing electrolyte to penetrate the material and conduct ions in the liquid phase to participate in the reactions. The active material in the anode is typically graphite and has an open-circuit potential close to zero, compared with lithium, and the active material in the cathode is commonly a transition metal or combination thereof, for example Nickel, Manganese and Cobalt or (NMC). The composition/chemistry of both electrodes, but especially the cathode, can vary quite significantly for different applications. 
Some materials, e.g., lithium-iron-phosphate (LFP), are more stable and longer lasting, but have a lower open-circuit voltage, and therefore, the resulting battery has lower power density.

Aside from chemistry, the dimensions of the layers and microstructural design of the materials also greatly affects whether a cell performs better as a high-energy or high-power cell. The losses within the cells are characterised by overpotentials that manifest by reducing the cell voltage, and these are typically dependent on the operating current. As batteries are usually operated within determined voltage windows for safety and longevity, the recoverable capacity (and cell voltage) decreases when cycling at higher rates. Two important transport related phenomena within the cell in particular are: ion  migration within the electrolyte, and diffusion within the active material. The interlayer transport is largely determined by the volume fractions of the electrode constituents, which also in turn affect the tortuosity of the transport pathways. The porosity of the solid electrode matrix refers to the volume fraction of the space that is filled with electrolyte. 
A decreased porosity here results in a corresponding increase in active material volume (and therefore specific capacity); however, the transport is then on average more difficult, and the cell will not be able to operate at high power. Similarly, if the electrodes are made thicker then the overall active material volume fraction will increase compared to inactive components such as current collectors and separators, but transport losses will increase as ions have further to travel to reach all parts of the cell. Finally, another key microstructural parameter is the average size of active particles. Smaller particles will have better solid diffusion losses and a greater active surface area per volume, but may necessitate a greater binder volume fraction in order to keep them structurally stable and provide enough electron pathways to the increased number of active sites.

Given all the above considerations, this case study seeks to select a set of hyper-parameters using
    published parameter sets from the battery literature
    \citep{Ai2019, Chen2020a, Ecker2015, Ecker2015a, Marquis2019, Yang2017} and then explore
    various geometrical parameters in order to find an optimal high energy and high power combination.
Table~\ref{tab:battery_notation} lists the notation for this case study.
\review{The mean power and discharged energy of the battery cell form the objective functions, i.e.
    $f_1$ and $f_2$, respectively, and are defined as:}

\begin{equation}
    \begin{cases}
        \review{f_1(\bm{x}) = \overline{IV},} \\
        \review{f_2(\bm{x}) = \int_{0}^{t} IV \,dt},
    \end{cases}
\end{equation}

\review{where $I$ is the current, $V$ is the voltage and $t$ is discharge time.
As simulations are performed on a 1-D domain an arbitrary cross-sectional area is assigned to the
    battery and determines the cell volume along with the layer thickness which is varied between cases.
Cell volume is used to normalize the objective functions making them comparable for different runs.}

\subsubsection{Optimization Benchmark}
% FIGURE: battery results
\begin{figure*}
    \begin{center}
        \includegraphics[width=0.78\paperwidth]{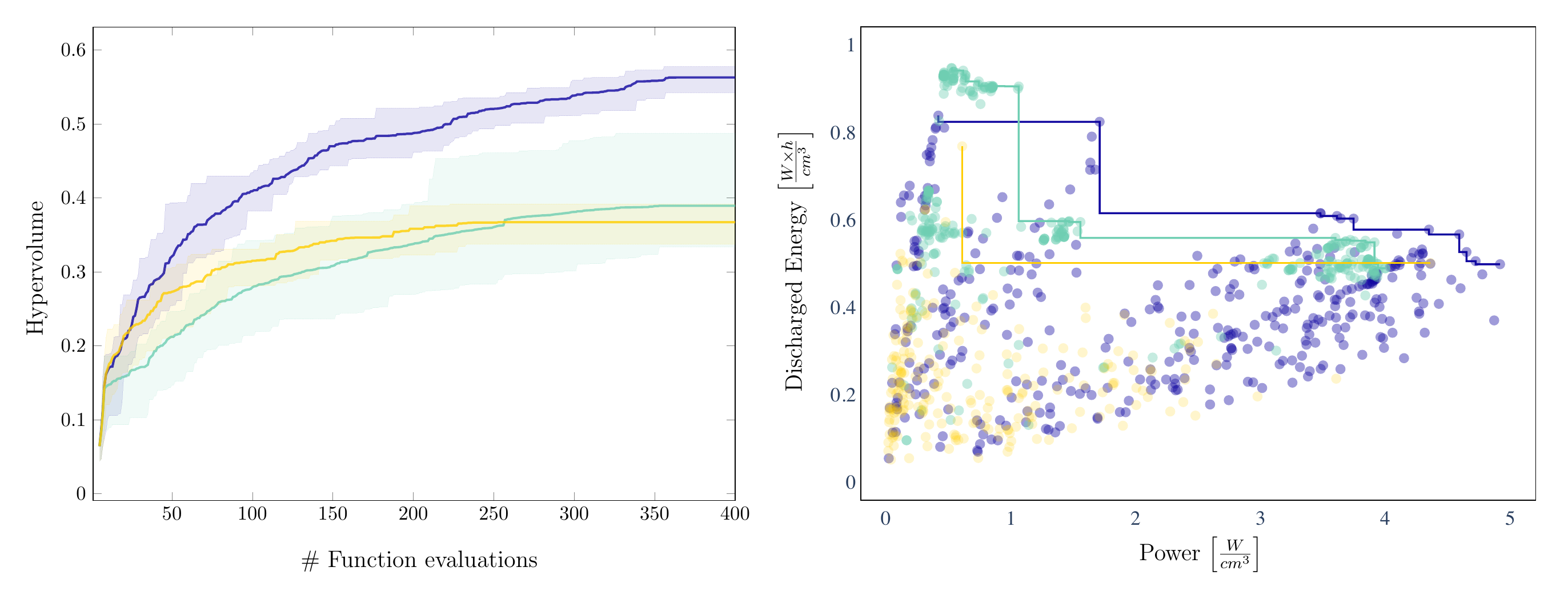}
    \end{center}
    \caption{Compares battery optimization results of \texttt{NSGA-II} (green),
    \texttt{ENTMOOT} (purple) and feasible sampling strategy (yellow). \textbf{(left)}~Hypervolume
    bounded by Pareto frontier approximation after a certain number of function evaluations. 
    \textbf{(right)}~Best Pareto frontier approximation found for different algorithms based on 
    hypervolume.}
    \label{fig:res_battery}
\end{figure*}
\review{Reasonable bounds for all optimization variables are given in
    Equ.~\eqref{eq:bat_h}~-~Equ.~\eqref{eq:bat_q}, with variable $p$ indicating the parameter
    set selected.
We model the variable $p$ as a categorical variable, as it is difficult to quantify a distance
    between different parameter sets $\{\text{Ai2019},\ \text{Chen2020},\ \text{Ecker2015},\
    \text{Marquis2019},\ \text{Yang2017}\}$.
}
Variables $\epsilon_{\text{poros}}$, $\epsilon_{\text{active}}$ and $R_{\text{particle}}$ denote
    porosity, active material volume fraction and particle radius
    and are optimized for both positive and negative side of the battery, i.e.\ indicated 
    by $+$ and $-$.
The negative and positive side electrode thickness is scaled by optimization variable $\lambda$ 
    allowing a continuous range between half and double the size of the default electrode.
Since porosity translates to the volume fraction of the electrolyte, Equ.~\eqref{eq:bat_a} and 
    Equ.~\eqref{eq:bat_b} limit the sum of porosity and active material volume fraction 
    $\epsilon_\text{active}$ to $0.95$ leaving a volume fraction of $5 \%$ for binder material.
Different battery specifications are designed for certain C-rate ranges, i.e.\ $1 \text{C}$ 
    translates to the current that discharges the battery in 1 hour, and is denoted by variable $C$.
Too high values of $C$ certain battery designs cause PyBaMM simulations to fail.
To reduce the number of failed simulations we introduce custom 
    upper bounds for $C$ in Equ.~\eqref{eq:bat_c}~-~Equ.\eqref{eq:bat_g}
    based on preliminary studies for all parameter sets in $p \in \{\text{Ai2019},\ 
    \text{Chen2020},\ \text{Ecker2015},\ \text{Marquis2019},\ \text{Yang2017}\} $ without modifying
    other optimization variables. 
Both objectives, i.e. \ power and discharged energy, are normalized based on best minimum and maximum values observed throughout
     all computational runs.
\review{In summary, this case study has one categorical variable describing the underlying
    parameter set that is being used, i.e.\ Equ.~\eqref{eq:bat_h}, and nine continuous variables
    denoting various design parameters of the battery cell that can be optimized, i.e.\
    Equ.~\eqref{eq:bat_i}~-~Equ.~\eqref{eq:bat_q}.
Equ.~\eqref{eq:bat_a}~-~Equ.~\eqref{eq:bat_g} comprise seven additional constraints for input variables.}

% added constraints
\textbf{Required Constraints}
\begin{subequations}
    \label{eq:req_constr_battery}
    \begin{flalign}
        & \epsilon_{\text{poros}}^{-} + \epsilon_{\text{active}}^{-} \leq 0.95 & & \label{eq:bat_a} \\
        & \epsilon_{\text{poros}}^{+} + \epsilon_{\text{active}}^{+} \leq 0.95 & & \label{eq:bat_b}\\
        & (p = \text{Ai2019}) \rightarrow C \leq 3.2 & & \label{eq:bat_c} \\
        & (p = \text{Chen2020}) \rightarrow C \leq 2.2 & & \label{eq:bat_d} \\
        & (p = \text{Ecker2015}) \rightarrow C \leq 8.2 & & \label{eq:bat_e} \\
        & (p = \text{Marquis2019}) \rightarrow C \leq 5.2 & & \label{eq:bat_f} \\
        & (p = \text{Yang2017}) \rightarrow C \leq 8.2 & & \label{eq:bat_g} \\
        & p \in \{\text{Ai2019},\ \text{Chen2020},\ \text{Ecker2015},\ 
        \text{Marquis2019},\ \text{Yang2017}\} 
        & & \label{eq:bat_h}\\
        & C \in \left[0.5, 8.2\right] & & \label{eq:bat_i} \\
        & \epsilon_{\text{poros}}^{-} \in \left[0.2,0.7 \right] & & \label{eq:bat_j} \\
        & \epsilon_{\text{active}}^{-} \in \left[0.2,0.7 \right] & & \label{eq:bat_k} \\
        & R_{\text{particle}}^{-} \in \left[\num{1e-6}, \num{20e-6} [m]\right] & & \label{eq:bat_l} \\
        & \epsilon_{\text{poros}}^{+} \in \left[0.2,0.7 \right] & & \label{eq:bat_m} \\
        & \epsilon_{\text{active}}^{+} \in \left[0.2,0.7 \right] & & \label{eq:bat_n} \\
        & R_{\text{particle}}^{+} \in \left[\num{1e-6}, \num{20e-6} [m]\right] & & \label{eq:bat_o} \\
        & \lambda^- \in \left[0.5,2.0\right] & & \label{eq:bat_p} \\
        & \lambda^+ \in \left[0.5,2.0\right] & & \label{eq:bat_q}
    \end{flalign}
\end{subequations}
\review{A procedure similar to Algorithm~\ref{algo:init} provides ten feasible initial points to all competing
    methods.
We fix the categorical variable $p \in \{\text{Ai2019},\ \text{Chen2020},\ \text{Ecker2015},\
    \text{Marquis2019},\ \text{Yang2017}\} $ for individual points to obtain two points for each
    possible parameter set $p$.
For the general case, the method extends Equ.~\eqref{eq:init} with similarity measures, e.g.\
    Equ.~\eqref{eq:overlap} and Equ.~\eqref{eq:goodall4}, to increase diversity in initial samples
    with respect to categorical features.} \\
Fig.~\ref{fig:res_battery} shows hypervolume improvement and best Pareto frontier approximations
    for the battery design benchmark.
All methods are capable improving the hypervolume of the objective space.
Compared to the Section~\ref{sec:wind} benchmark the constraints given in 
    Equ.~\eqref{eq:req_constr_battery} are less restrictive making feasible solutions more 
    attainable and allowing \texttt{NSGA-II} to surpass the feasible sampling strategy.
\texttt{ENTMOOT} manages to outperform other strategies for all different random seeds considered.
The Fig.~\ref{fig:res_battery} best Pareto frontier approximations derived by competing methods
    show that \texttt{NSGA-II} spends a large amount of evaluation budget to improve already explored
    points while \texttt{ENTMOOT} is actively exploring the entire objective space.
\texttt{ENTMOOT} especially dominates when it comes to high power solutions which are dominated by
    high C-rates.
Since, \texttt{ENTMOOT} guarantees feasibility of proposed solutions, C-rates are kept below the 
    bounds specified in Equ.~\eqref{eq:bat_c}~-~Equ.\eqref{eq:bat_g} allowing for a higher 
    frequency of successful simulations.
The feasible sampling strategy manages to only find two non-dominated points for the Pareto
    frontier approximation.

\section{Conclusion}
Energy systems with multiple opposing objects are challenging to optimize due to complex system behavior, highly restrictive 
    input constraints and large feature spaces with heterogenous variable types. 
The resulting problems are often handled using multi-objective Bayesian optimization, owing to relatively high sampling efficiency and ability to directly include input constraints. 
Tree-based models represent promising candidates for this class of optimization problems, as they are well-suited for nonlinear/discontinuous functions and naturally support discrete/categorical features. 
Nevertheless, uncertainty quantification and optimization over tree models are non-trivial, limiting their deployment in black-box optimization.

Given the above, this work introduces a multi-objective black-box optimization framework based on the tree-based \texttt{ENTMOOT} tool. 
The proposed all-in-one strategy addresses
    all of the above challenges, while also providing better sampling efficiency compared to the popular state-of-the-art tool \texttt{NGSA-II}.
\review{In a comprehensive numerical study, we demonstrate the advantages of the proposed framework on both synthetic
    problems and energy-related benchmarks, including windfarm layout optimization and lithium-ion battery design
    with two objective functions.}
In both energy-related benchmarks, \texttt{ENTMOOT} identifies better Pareto frontiers, and in significantly fewer
    iterations, compared to \texttt{NGSA-II} and a random search strategy.
Due to its versatility and sampling efficiency we envision \texttt{ENTMOOT} to be strongly relevant to real-world and industrial settings where experimental black-box evaluations are expensive.
\review{While the numerical studies only consider bi-objective problems,
    the proposed method also applies to case studies with more than two objective functions.
Future research will focus on other applications with more objective functions that can benefit from the proposed method.}

\clearpage

\section{Acknowledgments}
This work was supported by BASF SE, Ludwigshafen am Rhein, EPSRC Research Fellowships to RM (EP/P016871/1) and CT (EP/T001577/1), and an Imperial College Research Fellowship to CT. TT acknowledges funding from the EPSRC Faraday Institution Multiscale Modelling Project (EP/S003053/1, FIRG003)
\printbibliography
\end{document}